\documentclass[10pt,twocolumn,letterpaper]{article}

\usepackage{iccv}
\usepackage{times}
\usepackage{epsfig}
\usepackage{graphicx}
\usepackage{amsmath}
\usepackage{amssymb}

\usepackage{times}
\usepackage{epsfig}
\usepackage{graphicx}
\usepackage{float}
\usepackage{wrapfig}
\usepackage{amsmath,amssymb,amsthm}
\usepackage{algorithm,algorithmicx,algpseudocode}
\usepackage{bm,xspace}
\usepackage{comment}
\usepackage{verbatim}
\usepackage{multirow}
\usepackage{balance}
\usepackage{url}
\usepackage{booktabs}
\usepackage{etoolbox,siunitx}
\usepackage{calc}
\usepackage{pifont,hologo}

\usepackage[super]{nth}
\usepackage{nicefrac}
\robustify\bfseries

\def\pp{\mathbf{p}}

\def\xx{\mathbf{x}}

\def\pP{\mathcal{P}}

\def\xX{\mathcal{X}}

\DeclareMathSymbol{@}{\mathord}{letters}{"3B}

\newcommand\mypara[1]{\vspace{1mm}\noindent\textbf{#1}}

\def\latex/{\LaTeX}
\def\bibtex/{\hologo{BibTeX}}

\usepackage{enumitem}
\usepackage[accsupp]{axessibility}

\newcommand{\ve}[1]{\mathbf{#1}} 

\usepackage[pagebackref=true,breaklinks=true,letterpaper=true,colorlinks,bookmarks=false]{hyperref}

\iccvfinalcopy 

\def\iccvPaperID{****} 

\ificcvfinal\fi

\begin{document}

\title{Point Transformer}

\author{Hengshuang Zhao$^{1,2}$ \quad Li Jiang$^{3}$ \quad Jiaya Jia$^{3}$ \quad Philip Torr$^{1}$ \quad Vladlen Koltun$^{4}$\\
	$^{1}$University of Oxford \quad
	$^{2}$The University of Hong Kong \\
	$^{3}$The Chinese University of Hong Kong \quad
	$^{4}$Intel Labs\\ 
}

\maketitle
\ificcvfinal\thispagestyle{empty}\fi

\begin{abstract}
Self-attention networks have revolutionized natural language processing and are making impressive strides in image analysis tasks such as image classification and object detection. Inspired by this success, we investigate the application of self-attention networks to 3D point cloud processing. We design self-attention layers for point clouds and use these to construct self-attention networks for tasks such as semantic scene segmentation, object part segmentation, and object classification. Our Point Transformer design improves upon prior work across domains and tasks. For example, on the challenging S3DIS dataset for large-scale semantic scene segmentation, the Point Transformer attains an mIoU of 70.4\% on Area 5, outperforming the strongest prior model by 3.3 absolute percentage points and crossing the 70\% mIoU threshold for the first time.
\end{abstract}

\section{Introduction}
\label{sec:introduction}
3D data arises in many application areas such as autonomous driving, augmented reality, and robotics. Unlike images, which are arranged on regular pixel grids, 3D point clouds are sets embedded in continuous space. This makes 3D point clouds structurally different from images and precludes immediate application of deep network designs that have become standard in computer vision, such as networks based on the discrete convolution operator.

A variety of approaches to deep learning on 3D point clouds have arisen in response to this challenge. Some voxelize the 3D space to enable the application of 3D discrete convolutions~\cite{maturana2015voxnet,song2016ssc}. This induces massive computational and memory costs and underutilizes the sparsity of point sets in 3D. Sparse convolutional networks relieve these limitations by operating only on voxels that are not empty~\cite{graham20183d,choy20194d}.
Other designs operate directly on points and propagate information via pooling operators~\cite{qi2017pointnet,qi2017pointnet2} or continuous convolutions~\cite{wang2018pccn,thomas2019kpconv}. Another family of approaches connect the point set into a graph for message passing~\cite{wang2019dgcnn,li2019deepgcns}.

\begin{figure}
	\begin{center}
		\includegraphics[width=0.99\linewidth]{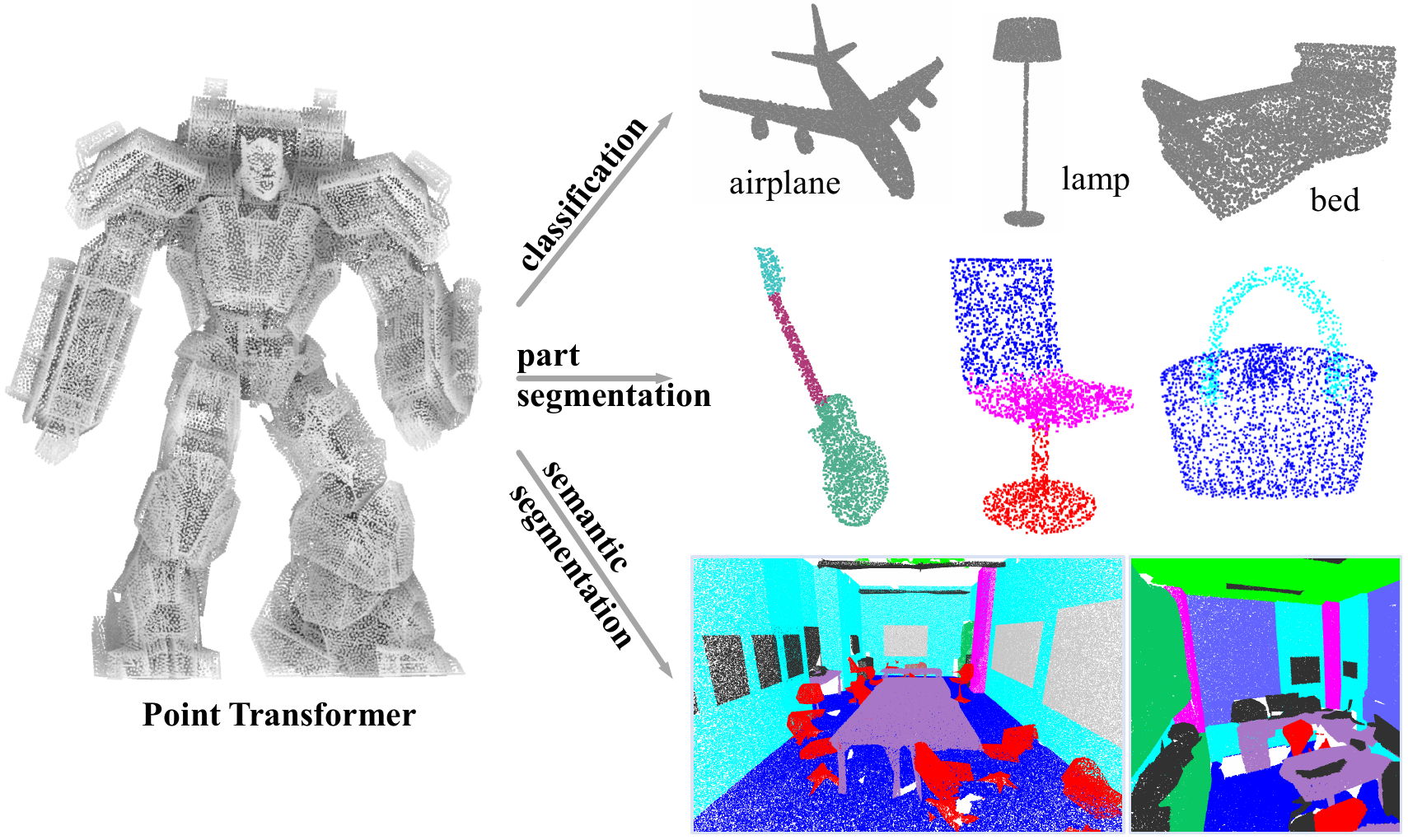}
	\end{center}
	\caption{The Point Transformer can serve as the backbone for various 3D point cloud understanding tasks such as object classification, object part segmentation, and semantic scene segmentation.}
	\label{fig:application}
\end{figure}

In this work, we develop an approach to deep learning on point clouds that is inspired by the success of transformers in natural language processing~\cite{vaswani2017attention,wu2019pay,Devlin2018,dai2019transformer,Yang2019xlnet} and image analysis~\cite{hu2019local,ramachandran2019stand,zhao2020san}. The transformer family of models is particularly appropriate for point cloud processing because the self-attention operator, which is at the core of transformer networks, is in essence a set operator: it is invariant to permutation and cardinality of the input elements. The application of self-attention to 3D point clouds is therefore quite natural, since point clouds are essentially sets embedded in 3D space.

We flesh out this intuition and develop a self-attention layer for 3D point cloud processing. Based on this layer, we construct Point Transformer networks for a variety of 3D understanding tasks. We investigate the form of the self-attention operator, the application of self-attention to local neighborhoods around each point, and the encoding of positional information in the network. The resulting networks are based purely on self-attention and pointwise operations.

We show that Point Transformers are remarkably effective in 3D deep learning tasks, both at the level of detailed object analysis and large-scale parsing of massive scenes. In particular, Point Transformers set the new state of the art on large-scale semantic segmentation on the S3DIS dataset (70.4\% mIoU on Area 5), shape classification on ModelNet40 (93.7\% overall accuracy), and object part segmentation on ShapeNetPart (86.6\% instance mIoU). Our full implementation and trained models will be released upon acceptance. In summary, our main contributions include the following.
\begin{itemize}[itemsep=0mm, topsep=2mm]
	\item
	We design a highly expressive Point Transformer layer for point cloud processing. The layer is invariant to permutation and cardinality and is thus inherently suited to point cloud processing.
	\item
	Based on the Point Transformer layer, we construct high-performing Point Transformer networks for classification and dense prediction on point clouds. These networks can serve as general backbones for 3D scene understanding.
	\item
	We report extensive experiments over multiple domains and datasets. We conduct controlled studies to examine specific choices in the Point Transformer design and set the new state of the art on multiple highly competitive benchmarks, outperforming long lines of prior work.
\end{itemize}

\section{Related Work}
\label{sec:related}
For 2D image understanding, pixels are placed in regular grids and can be processed with classical convolution. In contrast, 3D point clouds are unordered and scattered in 3D space: they are essentially sets. Learning-based approaches to processing 3D point clouds can be classified into the following types: projection-based, voxel-based, and point-based networks.

\mypara{Projection-based networks.}
For processing irregular inputs like point clouds, an intuitive way is to transform irregular representations to regular ones. Considering the success of 2D CNNs, some approaches~\cite{su15mvcnn,li2016vehicle,chen2017multi,kanezaki2018rotationnet,lang2019pointpillars} adopt multi-view projection, where 3D point clouds are projected into various image planes. Then 2D CNNs are used to extract feature representations in these image planes, followed by multi-view feature fusion to form the final output representations. In a related approach, TangentConv~\cite{tatarchenko2018tangent} projects local surface geometry onto a tangent plane at every point, forming tangent images that can be processed by 2D convolution. However, this approach heavily relies on tangent estimation. In projection-based frameworks, the geometric information inside point clouds is collapsed during the projection stage. These approaches may also underutilize the sparsity of point clouds when forming dense pixel  grids on projection planes. The choice of projection planes may heavily influence recognition performance and occlusion in 3D may impede accuracy.

\mypara{Voxel-based networks.}
An alternative approach to transforming irregular point clouds to regular representations is 3D voxelization~\cite{maturana2015voxnet,song2016ssc}, followed by convolutions in 3D. When applied naively, this strategy can incur massive computation and memory costs due to the cubic growth in the number of voxels as a function of resolution. The solution is to take advantage of sparsity, as most voxels are usually unoccupied. For example, OctNet~\cite{riegler2017octnet} uses unbalanced octrees with hierarchical partitions. Approaches based on sparse convolutions, where the convolution kernel is only evaluated at occupied voxels, can further reduce computation and memory requirements~\cite{graham20183d,choy20194d}. These methods have demonstrated good accuracy but may still lose geometric detail due to quantization onto the voxel grid.

\mypara{Point-based networks.}
Rather than projecting or quantizing irregular point clouds onto regular grids in 2D or 3D, researchers have designed deep network structures that ingest point clouds directly, as sets embedded in continuous space.
PointNet~\cite{qi2017pointnet} utilizes permutation-invariant operators such as pointwise MLPs and pooling layers to aggregate features across a set. PointNet++~\cite{qi2017pointnet2} applies these ideas within a hierarchical spatial structure to increase sensitivity to local geometric layout.
Such models can benefit from efficient sampling of the point set, and a variety of sampling strategies have been developed~\cite{qi2017pointnet2,dovrat2019learning,wu2019pointconv,yang2019modeling,hu2020randla}.

A number of approaches connect the point set into a graph and conduct message passing on this graph. DGCNN~\cite{wang2019dgcnn} performs graph convolutions on kNN graphs. PointWeb~\cite{zhao2019pointweb} densely connects local neightborhoods.
ECC~\cite{simonovsky2017dynamic} uses dynamic edge-conditioned filters where convolution kernels are generated based on edges inside point clouds. SPG~\cite{landrieu2018spg} operates on a superpoint graph that represents contextual relationships. KCNet~\cite{shen2018mining} utilizes kernel correlation and graph pooling. Wang et al.~\cite{wang2018local} investigate the local spectral graph convolution. 
GACNet~\cite{wang2019graph} employs graph attention convolution and HPEIN~\cite{jiang2019hpein} builds a hierarchical point-edge interaction architecture. DeepGCNs~\cite{li2019deepgcns} explore the advantages of depth in graph convolutional networks for 3D scene understanding.

A number of methods are based on continuous convolutions that apply directly to the 3D point set, with no quantization. PCCN~\cite{wang2018pccn} represents convolutional kernels as MLPs. SpiderCNN~\cite{xu2018spidercnn} defines kernel weights as a family of polynomial functions. Spherical CNN~\cite{esteves2018learning} designs spherical convolution to address the problem of 3D rotation equivariance. PointConv~\cite{wu2019pointconv} and KPConv~\cite{thomas2019kpconv} construct convolution weights based on the input coordinates. InterpCNN~\cite{mao2019interpolated} utilizes coordinates to interpolate pointwise kernel weights. PointCNN~\cite{li2018pointcnn} proposes to reorder the input unordered point clouds with special operators. Ummenhofer et al.~\cite{Ummenhofer2020} apply continuous convolutions to learn particle-based fluid dynamics.

\mypara{Transformer and self-attention.}
Transformer and self-attention models have revolutionized machine translation and natural language processing~\cite{vaswani2017attention,wu2019pay,Devlin2018,dai2019transformer,Yang2019xlnet}.
This has inspired the development of self-attention networks for 2D image recognition~\cite{hu2019local,ramachandran2019stand,zhao2020san,dosovitskiy2021image}. Hu et al.~\cite{hu2019local} and Ramachandran et al.~\cite{ramachandran2019stand} apply scalar dot-product self-attention within local image patches. Zhao et al.~\cite{zhao2020san} develop a family of vector self-attention operators. Dosovitskiy et al.~\cite{dosovitskiy2021image} treat images as sequences of patches.

Our work is inspired by the findings that transformers and self-attention networks can match or even outperform convolutional networks on sequences and 2D images. 
Self-attention is of particular interest in our setting because it is intrinsically a set operator: positional information is provided as attributes of elements that are processed as a set~\cite{vaswani2017attention,zhao2020san}. Since 3D point clouds are essentially sets of points with positional attributes, the self-attention mechanism seems particularly suitable to this type of data. We thus develop a Point Transformer layer that applies self-attention to 3D point clouds.

There are a number of previous works~\cite{xie2018attentional,liu2019point2sequence,yang2019modeling,lee2019set} that utilize attention for point cloud analysis. They apply global attention on the whole point cloud, which introduces heavy computation and renders these approaches inapplicable to large-scale 3D scene understanding. They also utilize scalar dot-product attention, where different channels share the same aggregation weights. In contrast, we apply self-attention locally, which enables scalability to large scenes with millions of points, and we utilize vector attention, which we show to be important for achieving high accuracy. We also demonstrate the importance of appropriate position encoding in large-scale point cloud understanding, in contrast to prior approaches that omitted position information. Overall, we show that appropriately designed self-attention networks can scale to large and complex 3D scenes, and substantially advance the state of the art in large-scale point cloud understanding.

\section{Point Transformer}
\label{sec:method}
\vspace{-2pt}
We begin by briefly revisiting the general formulation of transformers and self-attention operators. Then we present the point transformer layer for 3D point cloud processing. Lastly, we present our network architecture for 3D scene understanding.

\subsection{Background}
Transformers and self-attention networks have revolutionized natural language processing~\cite{vaswani2017attention,wu2019pay,Devlin2018,dai2019transformer,Yang2019xlnet} and have demonstrated impressive results in 2D image analysis~\cite{hu2019local,ramachandran2019stand,zhao2020san,dosovitskiy2021image}. Self-attention operators can be classified into two types: scalar attention~\cite{vaswani2017attention} and vector attention~\cite{zhao2020san}.

Let $\xX = \{\xx_i\}_i$ be a set of feature vectors. The standard scalar dot-product attention layer can be represented as follows:
\begin{equation}\label{eq:scalar}
\ve{y}_{i} = \sum_{\xx_j \in \xX} \rho\big(\varphi(\ve{x}_i)^\top \psi(\ve{x}_j) + \delta\big) \alpha(\ve{x}_{j}),
\end{equation}
where $\ve{y}_{i}$ is the output feature. $\varphi$, $\psi$, and $\alpha$ are pointwise feature transformations, such as linear projections or MLPs. $\delta$ is a position encoding function and $\rho$ is a normalization function such as \textit{softmax}. The scalar attention layer computes the scalar product between features transformed by $\varphi$ and $\psi$ and uses the output as an attention weight for aggregating features transformed by $\alpha$.

In vector attention, the computation of attention weights is different. In particular, attention weights are \emph{vectors} that can modulate individual feature channels:
\begin{equation}\label{eq:vector}
\ve{y}_{i} = \sum_{\xx_j \in \xX} \rho\big(\gamma(\beta(\varphi(\ve{x}_i), \psi(\ve{x}_j))+\delta)\big) \odot \alpha(\ve{x}_{j}),
\end{equation}
where $\beta$ is a relation function (e.g., subtraction) and $\gamma$ is a mapping function (e.g., an MLP) that produces attention vectors for feature aggregation.

Both scalar and vector self-attention are set operators. The set can be a collection of feature vectors that represent the entire signal (e.g., sentence or image)~\cite{vaswani2017attention,dosovitskiy2021image} or a collection of feature vectors from a local patch within the signal (e.g., an image patch)~\cite{hu2019local,ramachandran2019stand,zhao2020san}.

\subsection{Point Transformer Layer}
\label{sec:layer}
\begin{figure}[h]
	\centering
	\includegraphics[width=0.98\linewidth]{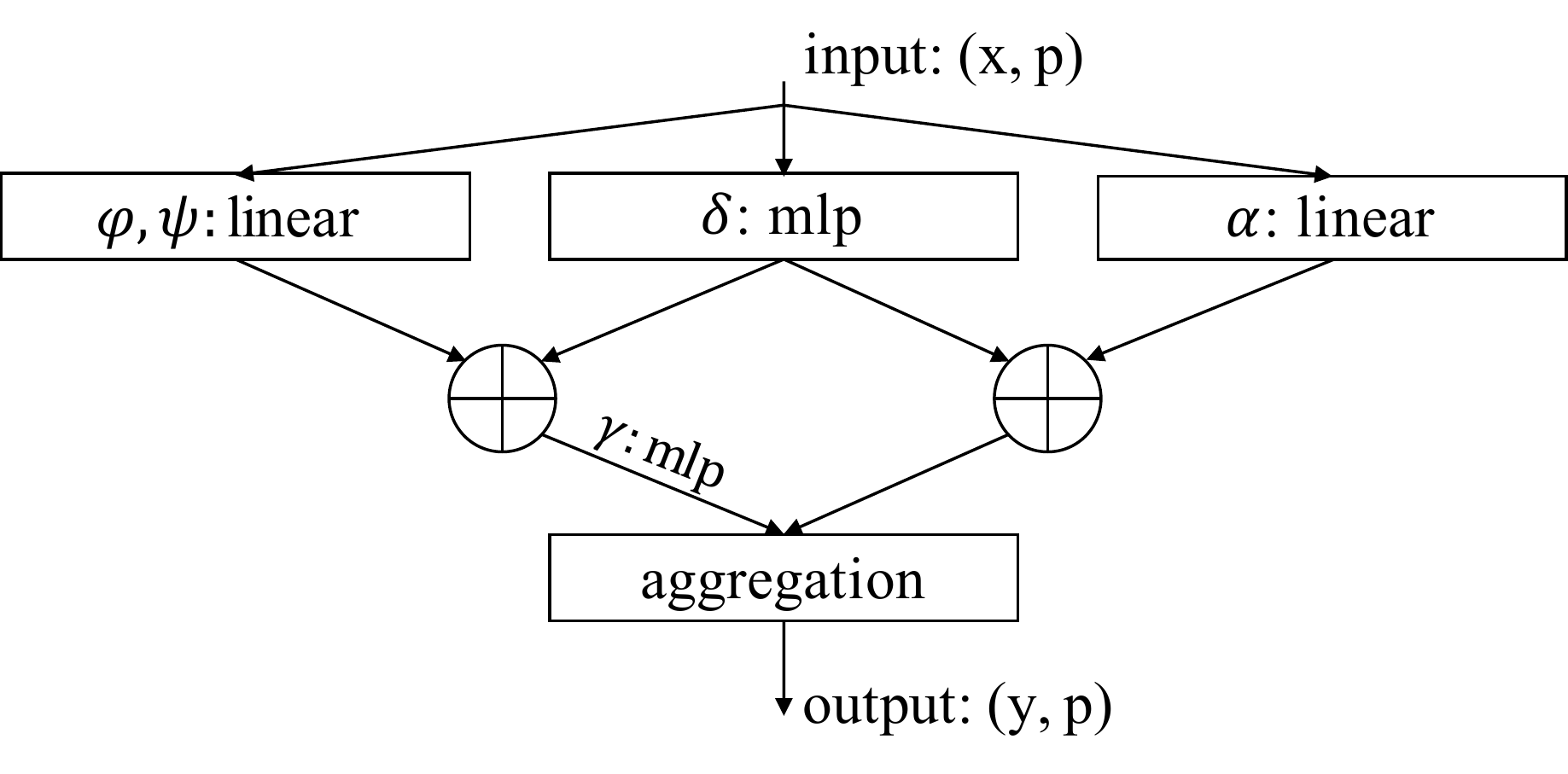} \\
	\caption{Point transformer layer.}
	\label{fig:transformerlayer}
\end{figure}

Self-attention is a natural fit for point clouds because point clouds are essentially sets embedded irregularly in a metric space. Our point transformer layer is based on vector self-attention. We use the subtraction relation and add a position encoding $\delta$ to both the attention vector $\gamma$ and the transformed features $\alpha$:
\begin{equation}\label{eq:pointtransformer}
\ve{y}_{i} = \sum_{\xx_j \in \xX(i)} \rho\big(\gamma(\varphi(\ve{x}_i) - \psi(\ve{x}_j) + \delta)\big) \odot \big(\alpha(\ve{x}_{j}) + \delta\big)
\end{equation}
Here the subset $\xX(i) \subseteq \xX$ is a set of points in a local neighborhood (specifically, $k$ nearest neighbors) of $\xx_i$. Thus we adopt the practice of recent self-attention networks for image analysis in applying self-attention locally, within a local neighborhood around each datapoint~\cite{hu2019local,ramachandran2019stand,zhao2020san}. The mapping function $\gamma$ is an MLP with two linear layers and one ReLU nonlinearity. The point transformer layer is illustrated in Figure~\ref{fig:transformerlayer}.

\begin{figure*}[t]
	\includegraphics[width=1.0\linewidth]{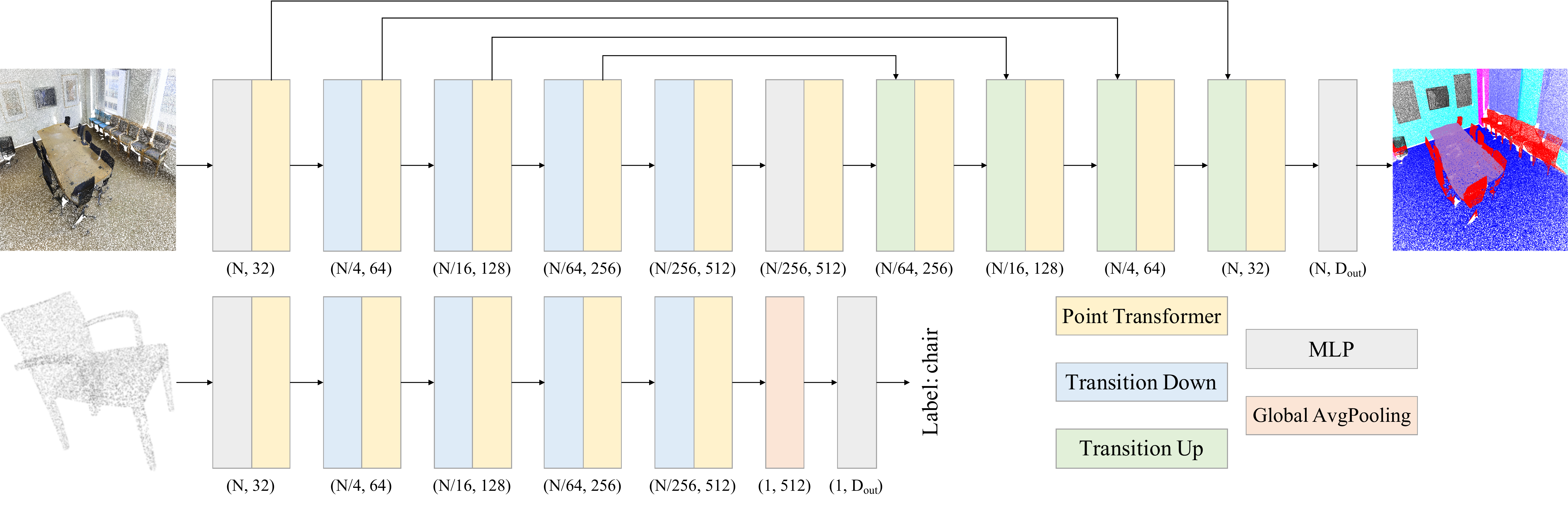}
	\caption{Point transformer networks for semantic segmentation (top) and classification (bottom).}
	\label{fig:network}
\end{figure*}

\begin{figure*}
	\centering
	\begin{tabular}{@{\hspace{8.0mm}}c@{\hspace{15.0mm}}c@{\hspace{8.0mm}}c@{\hspace{0.0mm}}}
		\includegraphics[height=0.24\linewidth]{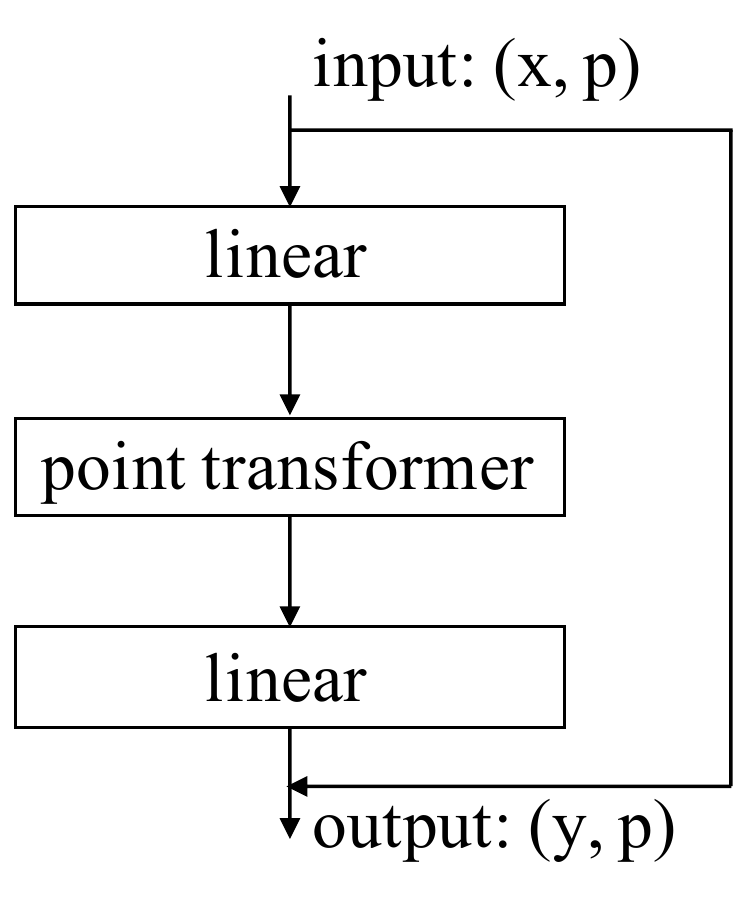} &
		\includegraphics[height=0.24\linewidth]{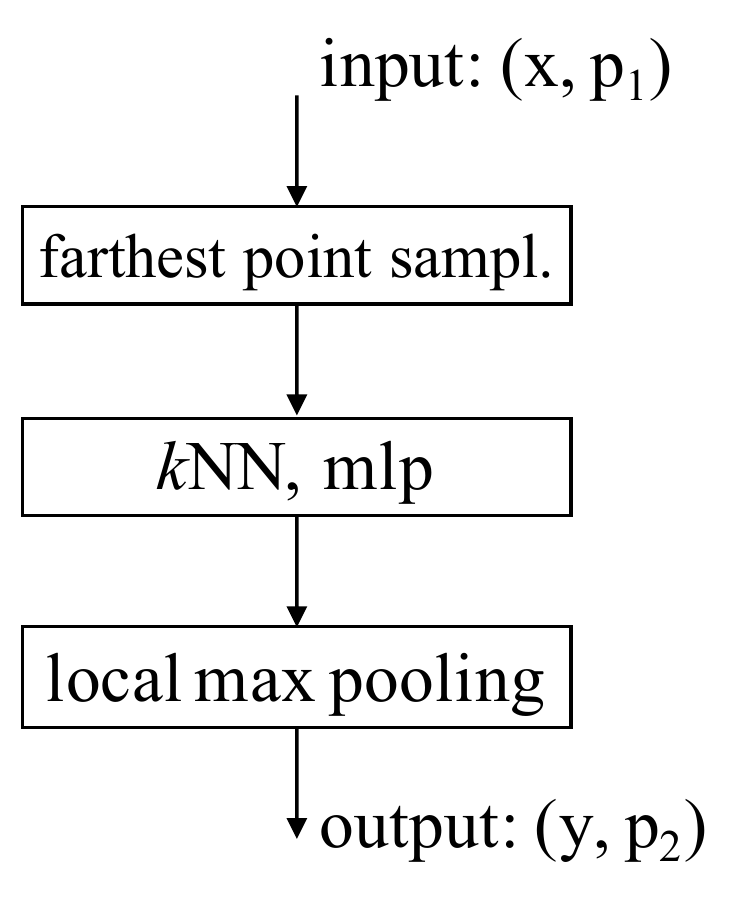} & \includegraphics[height=0.24\linewidth]{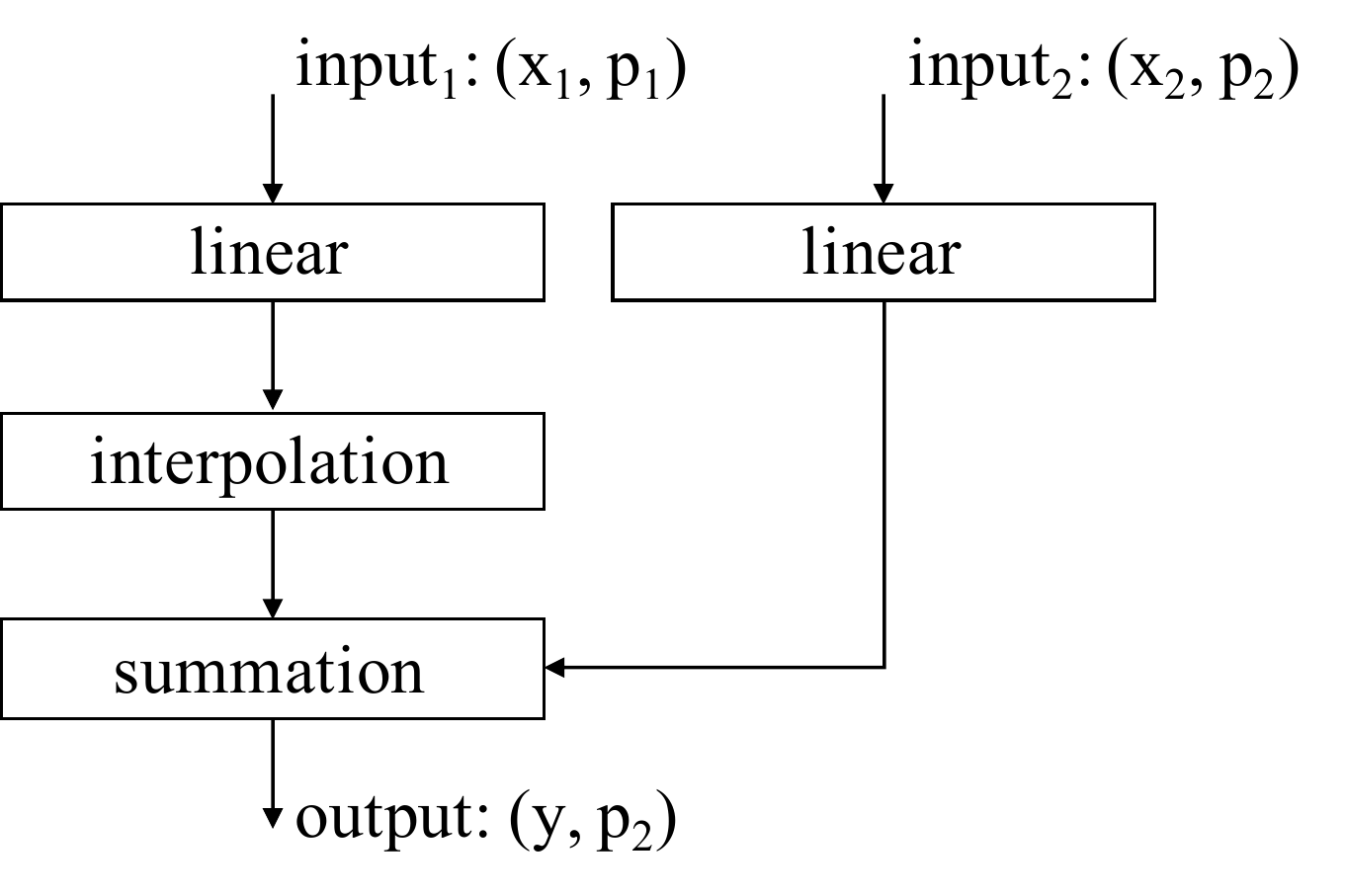} \\
	    \small (a) point transformer block & \small (b) transition down & \small (c) transition up \\
	\end{tabular}
	\vspace{4mm}
	\caption{Detailed structure design for each module.}
	\label{fig:detailedstructure}
\end{figure*}

\subsection{Position Encoding}
Position encoding plays an important role in self-attention, allowing the operator to adapt to local structure in the data~\cite{vaswani2017attention}. Standard position encoding schemes for sequences and image grids are crafted manually, for example based on sine and cosine functions or normalized range values~\cite{vaswani2017attention,zhao2020san}. In 3D point cloud processing, the 3D point coordinates themselves are a natural candidate for position encoding. We go beyond this by introducing trainable, parameterized position encoding. Our position encoding function $\delta$ is defined as follows:
\begin{equation}\label{eq:positionencoding}
\delta = \theta(\ve{p}_i - \ve{p}_j).
\end{equation}
Here $\ve{p}_i$ and $\ve{p}_j$ are the 3D point coordinates for points $i$ and $j$. The encoding function $\theta$ is an MLP with two linear layers and one ReLU nonlinearity. Notably, we found that position encoding is important for both the attention generation branch and the feature transformation branch. Thus Eq.~\ref{eq:pointtransformer} adds the trainable position encoding in both branches. The position encoding $\theta$ is trained end-to-end with the other subnetworks.

\subsection{Point Transformer Block}
We construct a residual point transformer block with the point transformer layer at its core, as shown in Figure~\ref{fig:detailedstructure}(a). The transformer block integrates the self-attention layer, linear projections that can reduce dimensionality and accelerate processing, and a residual connection. The input is a set of feature vectors $\xx$ with associated 3D coordinates $\pp$. The point transformer block facilitates information exchange between these localized feature vectors, producing new feature vectors for all data points as its output. The information aggregation adapts both to the content of the feature vectors and their layout in 3D.

\subsection{Network Architecture}
We construct complete 3D point cloud understanding networks based on the point transformer block. Note that the point transformer is the primary feature aggregation operator throughout the network. We do not use convolutions for preprocessing or auxiliary branches: the network is based entirely on point transformer layers, pointwise transformations, and pooling. The network architectures are visualized in Figure~\ref{fig:network}.

\mypara{Backbone structure.}
The feature encoder in point transformer networks for semantic segmentation and classification has five stages that operate on progressively downsampled point sets. The downsampling rates for the stages are [1, 4, 4, 4, 4], thus the cardinality of the point set produced by each stage is [N, N/4, N/16, N/64, N/256], where N is the number of input points. Note that the number of stages and the downsampling rates can be varied depending on the application, for example to construct light-weight backbones for fast processing. Consecutive stages are connected by transition modules: transition down for feature encoding and transition up for feature decoding.

\mypara{Transition down.}
A key function of the transition down module is to reduce the cardinality of the point set as required, for example from N to N/4 in the transition from the first to the second stage. Denote the point set provided as input to the transition down module as $\pP_1$ and denote the output point set as $\pP_2$. We perform farthest point sampling~\cite{qi2017pointnet2} in $\pP_1$ to identify a well-spread subset $\pP_2 \subset \pP_1$ with the requisite cardinality. To pool feature vectors from $\pP_1$ onto $\pP_2$, we use a $k$NN graph on $\pP_1$. (This is the same $k$ as in Section~\ref{sec:layer}. We use $k=16$ throughout and report a controlled study of this hyperparameter in Section~\ref{sec:ablation}.) Each input feature goes through a linear transformation, followed by batch normalization and ReLU, followed by max pooling onto each point in $\pP_2$ from its $k$ neighbors in $\pP_1$. The transition down module is schematically illustrated in Figure~\ref{fig:detailedstructure}(b).

\mypara{Transition up.}
For dense prediction tasks such as semantic segmentation, we adopt a U-net design in which the encoder described above is coupled with a symmetric decoder~\cite{qi2017pointnet2,choy20194d}. Consecutive stages in the decoder are connected by transition up modules. Their primary function is to map features from the downsampled input point set $\pP_2$ onto its superset $\pP_1 \supset \pP_2$.
To this end, each input point feature is processed by a linear layer, followed by batch normalization and ReLU, and then the features are mapped onto the higher-resolution point set $\pP_1$ via trilinear interpolation.
These interpolated features from the preceding decoder stage are summarized with the features from the corresponding encoder stage, provided via a skip connection.
The structure of the transition up module is illustrated in Figure~\ref{fig:detailedstructure}(c).

\mypara{Output head.}
For semantic segmentation, the final decoder stage produces a feature vector for each point in the input point set. We apply an MLP to map this feature to the final logits. For classification, we perform global average pooling over the pointwise features to get a global feature vector for the whole point set. This global feature is passed through an MLP to get the global classification logits.

\section{Experiments}
\label{sec:experiments}
\begin{table*}[!th]
	\centering
	\resizebox{1.0\linewidth}{!}{
		\begin{tabular}{ l | c c c | c c c c c c c c c c c c c}
			\toprule[1pt]
			Method & OA & mAcc & mIoU & ceiling & floor & wall & beam & column & window & door & table & chair & sofa & bookcase & board & clutter \\
			\hline
			PointNet~\cite{qi2017pointnet} & -- & 49.0 & 41.1 & 88.8 & 97.3 & 69.8 & 0.1 & 3.9 & 46.3 & 10.8 & 59.0 & 52.6 & 5.9 & 40.3 & 26.4 & 33.2\\
			SegCloud~\cite{tchapmi2017segcloud} & -- & 57.4 & 48.9 & 90.1 & 96.1 & 69.9 & 0.0 & 18.4 & 38.4 & 23.1 & 70.4 & 75.9 & 40.9 & 58.4 & 13.0 & 41.6\\
			TangentConv~\cite{tatarchenko2018tangent} & -- & 62.2 & 52.6 & 90.5 & 97.7 & 74.0 & 0.0 & 20.7 & 39.0 & 31.3 & 77.5 & 69.4 & 57.3 & 38.5 & 48.8 & 39.8 \\
			PointCNN~\cite{li2018pointcnn} & 85.9 & 63.9 & 57.3 & 92.3 & 98.2 & 79.4 & 0.0 & 17.6 & 22.8 & 62.1 & 74.4 & 80.6 & 31.7 & 66.7 & 62.1 & 56.7\\
			SPGraph~\cite{landrieu2018spg} & 86.4 & 66.5 & 58.0 & 89.4 & 96.9 & 78.1 & 0.0 & 42.8 & 48.9 & 61.6 & 84.7 & 75.4 & 69.8 & 52.6 & 2.1 & 52.2\\
			PCCN~\cite{wang2018pccn} & -- & 67.0 & 58.3 & 92.3 & 96.2 & 75.9 & 0.3 & 6.0 & 69.5 & 63.5 & 66.9 & 65.6 & 47.3 & 68.9 & 59.1 & 46.2\\
			PAT~\cite{yang2019modeling} & -- & 70.8 & 60.1 & 93.0 & 98.5 & 72.3 & 1.0 & 41.5 & 85.1 & 38.2 & 57.7 & 83.6 & 48.1 & 67.0 & 61.3 & 33.6\\
			PointWeb~\cite{zhao2019pointweb} & 87.0 & 66.6 & 60.3 & 92.0 & 98.5 & 79.4 & 0.0 & 21.1 & 59.7 & 34.8 & 76.3 & 88.3 & 46.9 & 69.3 & 64.9 & 52.5 \\
			HPEIN~\cite{jiang2019hpein} & 87.2 & 68.3 & 61.9 & 91.5 & 98.2 & 81.4 & 0.0 & 23.3 & 65.3 & 40.0 & 75.5 & 87.7 & 58.5 & 67.8 & 65.6 & 49.4 \\
			MinkowskiNet~\cite{thomas2019kpconv} & -- & 71.7 & 65.4 & 91.8 & 98.7 & 86.2 & 0.0 & 34.1 & 48.9 & 62.4 & 81.6 & 89.8 & 47.2 & 74.9 & 74.4 & 58.6 \\
			KPConv~\cite{thomas2019kpconv} & -- & 72.8 & 67.1 &92.8 & 97.3 & 82.4 & 0.0 & 23.9 & 58.0 & 69.0 & 81.5 & 91.0 & 75.4 & 75.3 & 66.7 & 58.9 \\
			\hline
			PointTransformer & \textbf{90.8} & \textbf{76.5} & \textbf{70.4} & 94.0 & 98.5 & 86.3 & 0.0 & 38.0 & 63.4 & 74.3 & 89.1 & 82.4 & 74.3 & 80.2 & 76.0 & 59.3 \\
			\bottomrule[1pt]
	\end{tabular}}
	\caption{Semantic segmentation results on the S3DIS dataset, evaluated on Area 5.}
	\label{tab:s3disresult}
\end{table*}

\begin{table}[!th]
	\centering
		\begin{tabular}{ l | c c c}
			\toprule[1pt]
			Method & OA & mAcc & mIoU \\
			\hline
			PointNet~\cite{qi2017pointnet} & 78.5 & 66.2 & 47.6 \\
			RSNet~\cite{huang2018recurrent} & -- & 66.5 & 56.5 \\
			SPGraph~\cite{landrieu2018spg} & 85.5 & 73.0 & 62.1 \\
			PAT~\cite{yang2019modeling} & -- & 76.5 & 64.3 \\
			PointCNN~\cite{li2018pointcnn} & 88.1 & 75.6 & 65.4 \\
			PointWeb~\cite{zhao2019pointweb} & 87.3 & 76.2 & 66.7 \\
			ShellNet~\cite{zhang2019shellnet} & 87.1 & -- & 66.8 \\
			RandLA-Net~\cite{thomas2019kpconv} & 88.0 & 82.0 & 70.0 \\
			KPConv~\cite{thomas2019kpconv} & -- & 79.1 & 70.6 \\
			\hline
			PointTransformer & \textbf{90.2} & \textbf{81.9} & \textbf{73.5} \\
			\bottomrule[1pt]
	\end{tabular}
	\caption{Semantic segmentation results on the S3DIS dataset, evaluated with 6-fold cross-validation.}
	\label{tab:s3disresult2}
\end{table}

We evaluate the effectiveness of the presented Point Transformer design on a number of domains and tasks. For 3D semantic segmentation, we use the challenging Stanford Large-Scale 3D Indoor Spaces (S3DIS) dataset~\cite{armeni2016s3dis}. For 3D shape classification, we use the widely adopted ModelNet40 dataset~\cite{wu2015modelnet}. And for object part segmentation, we use ShapeNetPart~\cite{yi2016scalable}.

\mypara{Implementation details.}
We implement the Point Transformer in PyTorch~\cite{Paszke2019Pytorch}. We use the SGD optimizer with momentum and weight decay set to 0.9 and 0.0001, respectively. For semantic segmentation on S3DIS, we train for 40K iterations with initial learning rate 0.5, dropped by 10x at steps 24K and 32K. For 3D shape classification on ModelNet40 and 3D object part segmentation on ShapeNetPart, we train for 200 epochs. The initial learning rate is set to 0.05 and is dropped by 10x at epochs 120 and 160.

\subsection{Semantic Segmentation}

\noindent\textbf{Data and metric.}
The S3DIS~\cite{armeni2016s3dis} dataset for semantic scene parsing consists of 271 rooms in six areas from three different buildings. Each point in the scan is assigned a semantic label from 13 categories (ceiling, floor, table, etc.). Following a common protocol~\cite{tchapmi2017segcloud,qi2017pointnet2}, we evaluate the presented approach in two modes: (a) Area 5 is withheld during training and is used for testing, and (b) 6-fold cross-validation. For evaluation metrics, we use mean classwise intersection over union (mIoU), mean of classwise accuracy (mAcc), and overall pointwise accuracy (OA).

\mypara{Performance comparison.}
The results are presented in Tables~\ref{tab:s3disresult} and~\ref{tab:s3disresult2}. The Point Transformer outperforms all prior models according to all metrics in both evaluation modes. On Area 5, the Point Transformer attains mIoU/mAcc/OA of 70.4\%/76.5\%/90.8\%, outperforming all prior work by multiple percentage points in each metric. The Point Transformer is the first model to pass the 70\% mIoU bar, outperforming the prior state of the art by 3.3 absolute percentage points in mIoU. The Point Transformer outperforms MLPs-based frameworks such as PointNet~\cite{qi2017pointnet}, voxel-based architectures such as SegCloud~\cite{tchapmi2017segcloud}, graph-based methods such as SPGraph~\cite{landrieu2018spg}, attention-based methods such as PAT~\cite{yang2019modeling}, sparse convolutional networks such as MinkowskiNet~\cite{choy20194d}, and continuous convolutional networks such as KPConv~\cite{thomas2019kpconv}.
Point Transformer also substantially outperforms all prior models under 6-fold cross-validation. The mIoU in this mode is 73.5\%, outperforming the prior state of the art (KPConv) by 2.9 absolute percentage points. The number of parameters in Point Transformer (4.9M) is much smaller than in current high-performing architectures such as KPConv (14.9M) and SparseConv (30.1M).

\begin{figure*}
	\centering
	\small
	\resizebox{0.97\linewidth}{!}{
	\begin{tabular}{@{\hspace{0.0mm}}c@{\hspace{1.0mm}}c@{\hspace{1.0mm}}c@{\hspace{2.5mm}}c@{\hspace{1.0mm}}c@{\hspace{1.0mm}}c@{\hspace{0.0mm}}}
		Input & Ground Truth & Point Transformer & Input & Ground Truth & Point Transformer \\
		\includegraphics[width=0.15\linewidth]{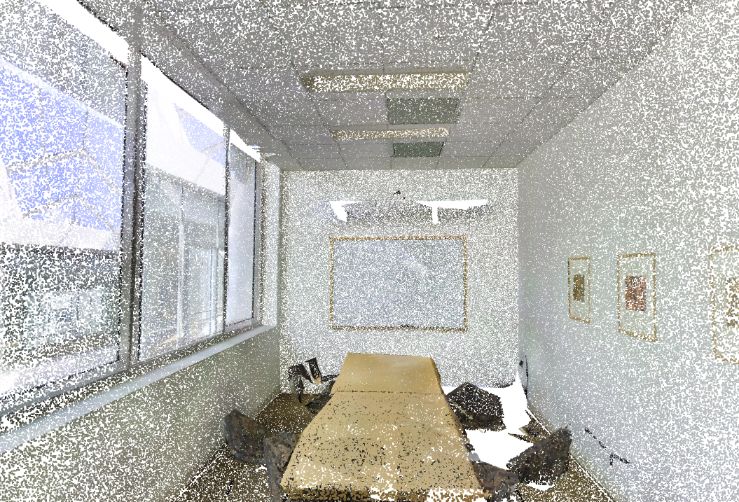}&
		\includegraphics[width=0.15\linewidth]{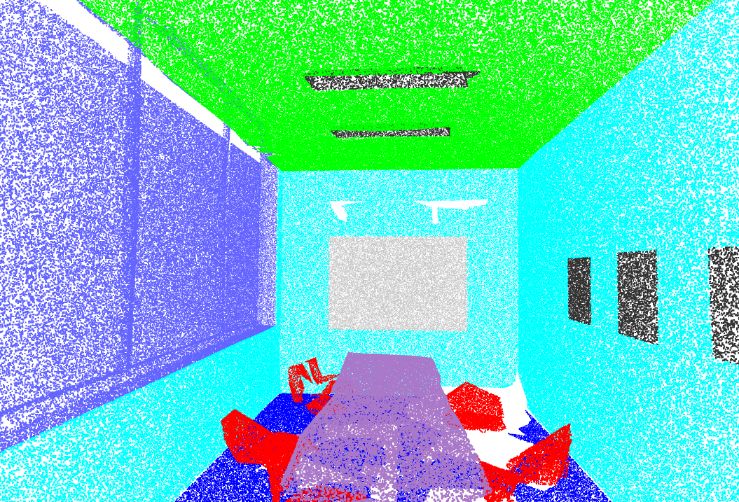}&
		\includegraphics[width=0.15\linewidth]{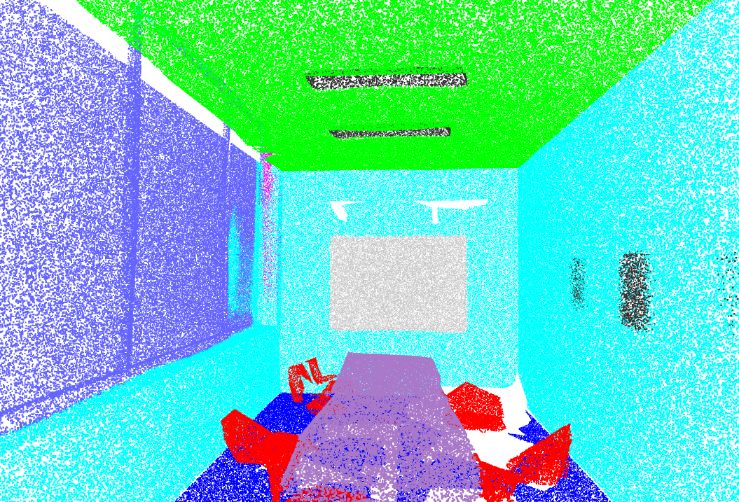}&
		\includegraphics[width=0.15\linewidth]{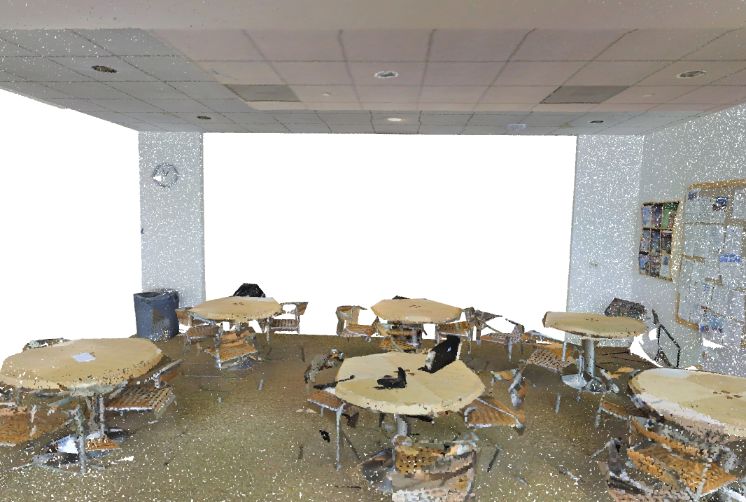}&
		\includegraphics[width=0.15\linewidth]{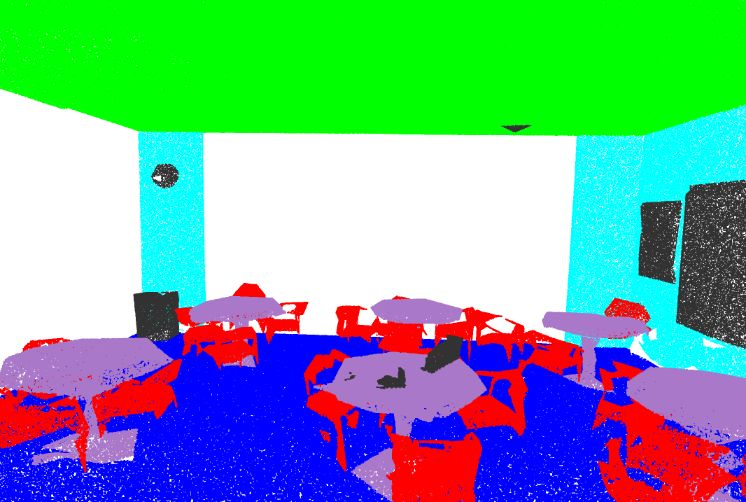}&
		\includegraphics[width=0.15\linewidth]{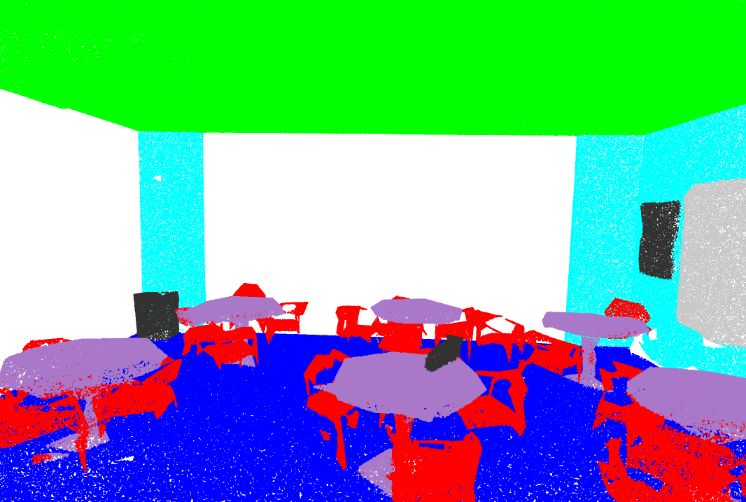}\\
		\includegraphics[width=0.15\linewidth]{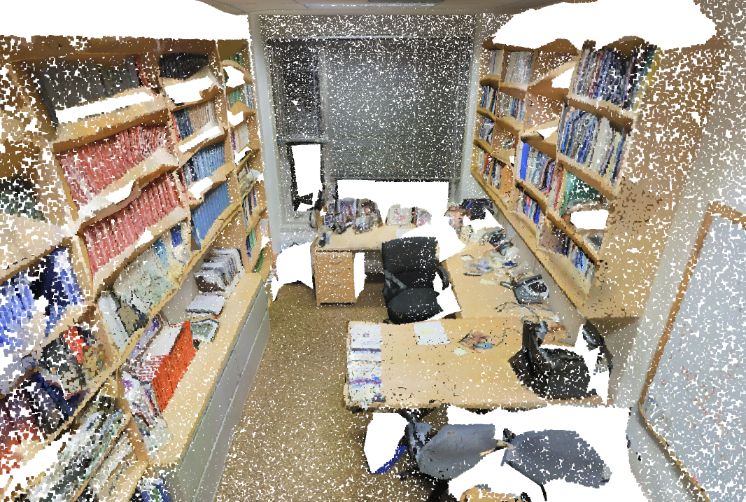}&
		\includegraphics[width=0.15\linewidth]{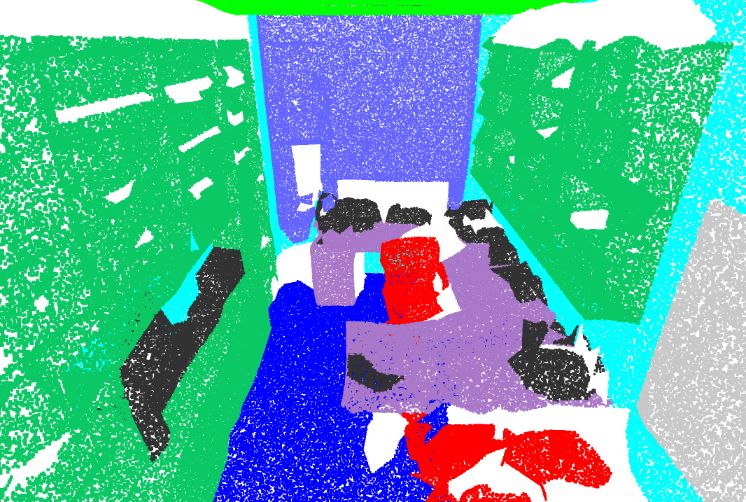}&
		\includegraphics[width=0.15\linewidth]{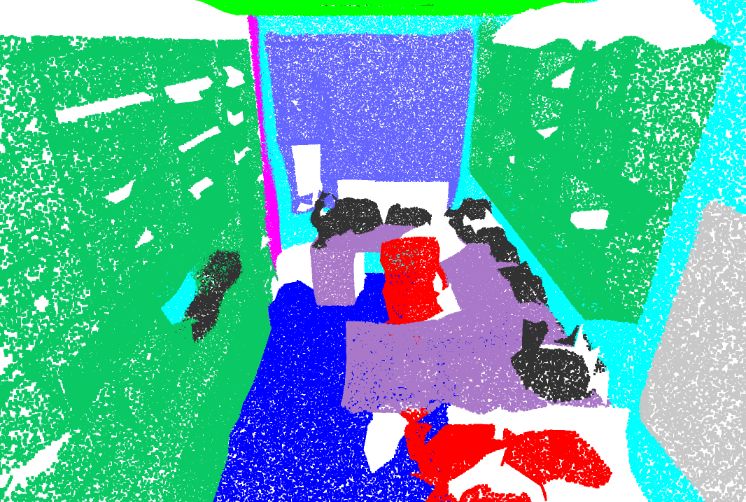}&
		\includegraphics[width=0.15\linewidth]{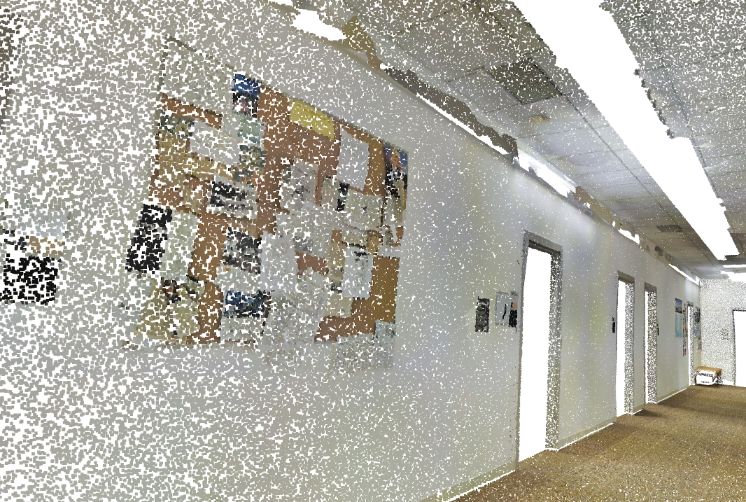}&
		\includegraphics[width=0.15\linewidth]{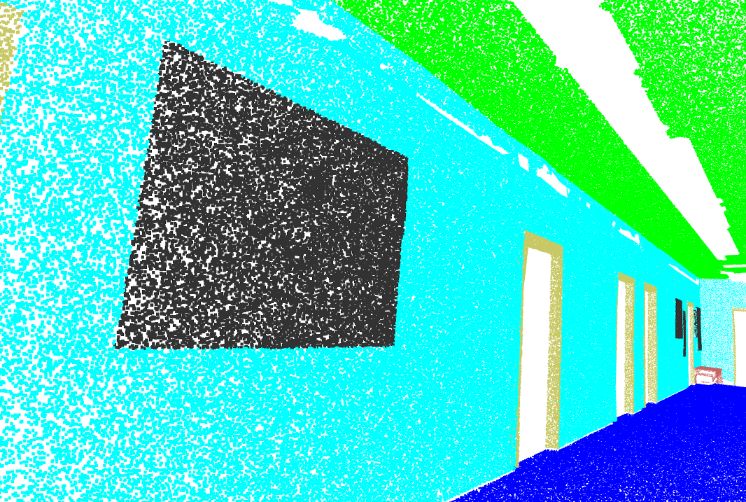}&
		\includegraphics[width=0.15\linewidth]{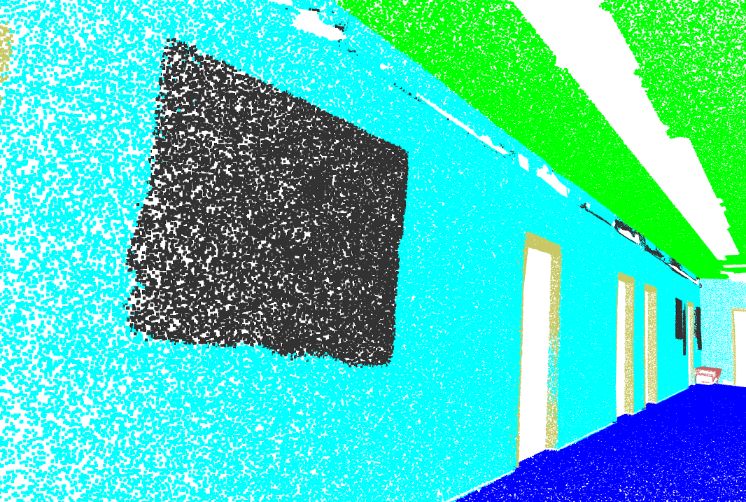}\\
		\includegraphics[width=0.15\linewidth]{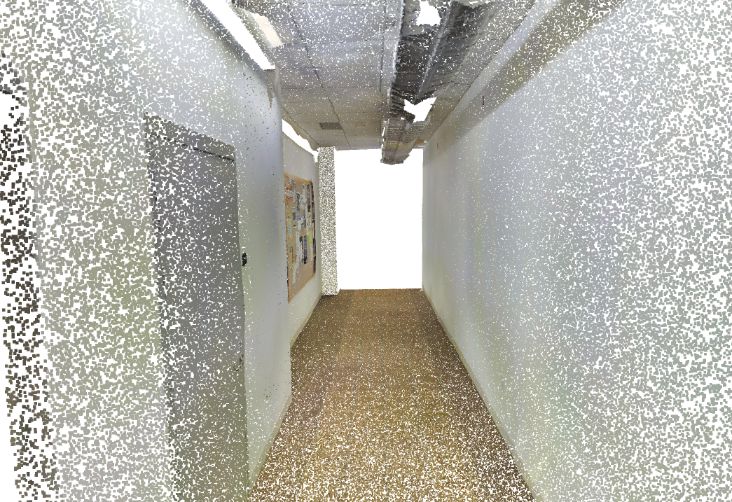}&
		\includegraphics[width=0.15\linewidth]{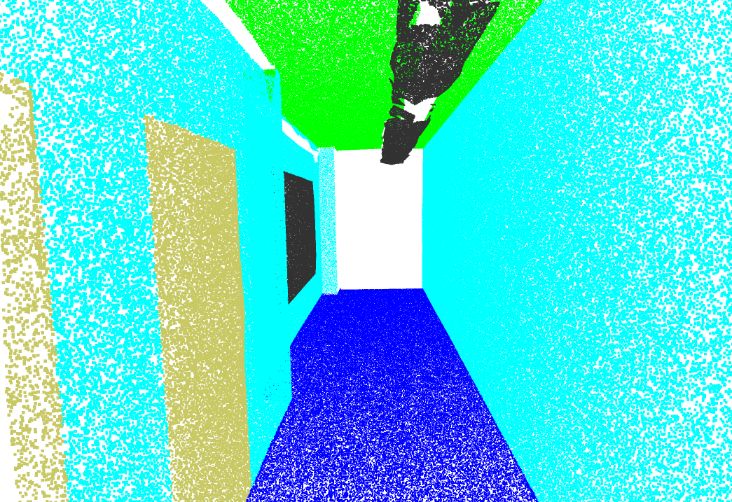}&
		\includegraphics[width=0.15\linewidth]{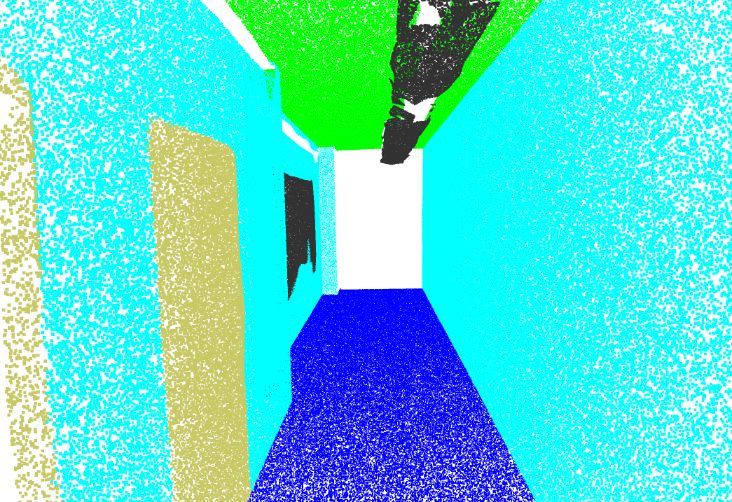}&
		\includegraphics[width=0.15\linewidth]{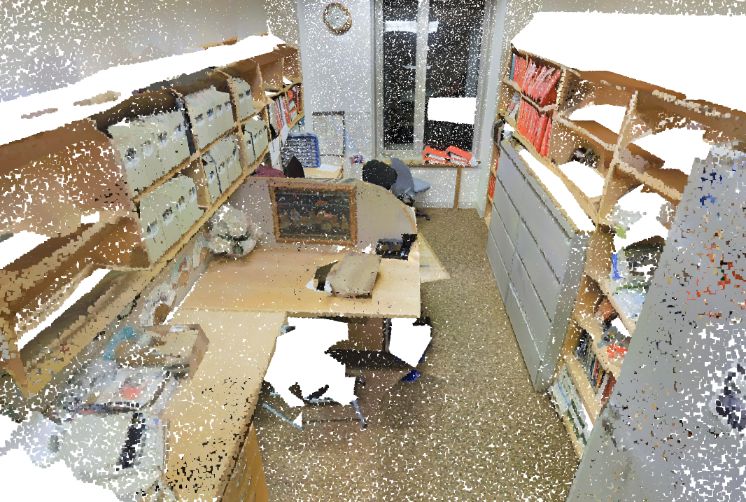}&
		\includegraphics[width=0.15\linewidth]{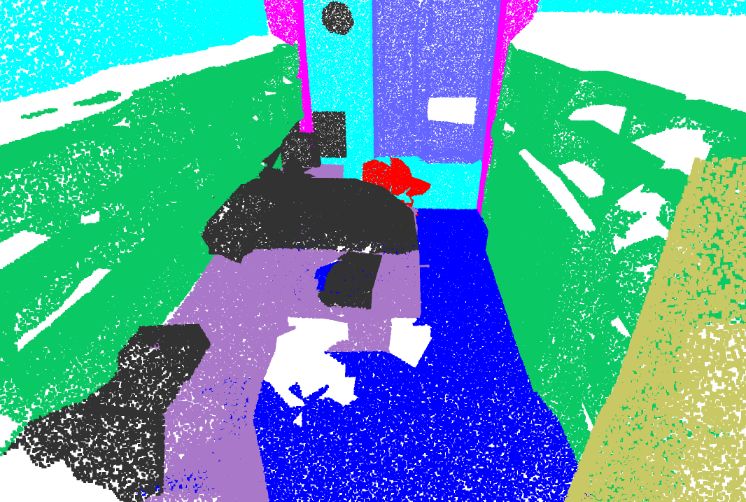}&
		\includegraphics[width=0.15\linewidth]{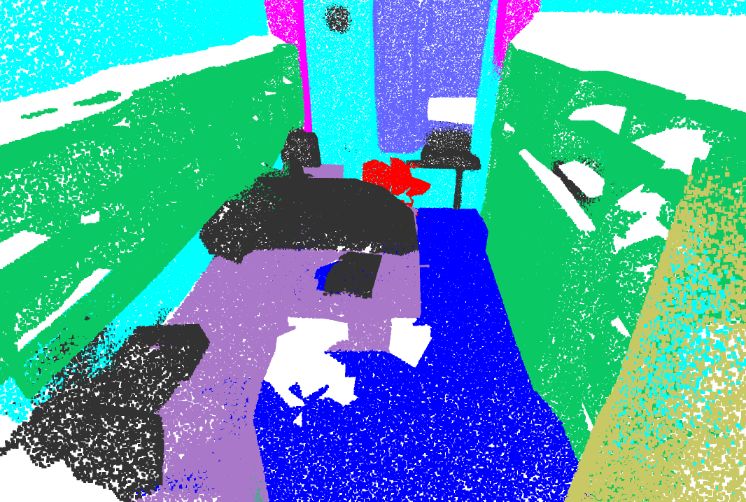}\\
		\multicolumn{6}{c}{\includegraphics[width=0.9\linewidth]{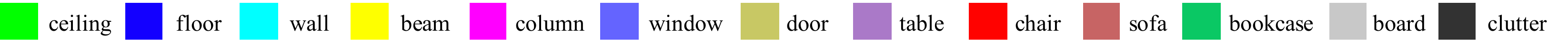}}\\
	\end{tabular}}
	\caption{Visualization of semantic segmentation results on the S3DIS dataset.}
	\label{fig:s3disvisual}
\end{figure*}

\mypara{Visualization.}
Figure~\ref{fig:s3disvisual} shows the Point Transformer's predictions.
We can see that the predictions are very close to the ground truth. Point Transformer captures detailed semantic structure in complex 3D scenes, such as the legs of chairs, the outlines of poster boards, and the trim around doorways.

\begin{table}[h]
	\centering
	\begin{tabular}{ l | c c c}
		\toprule[1pt]
		Method & input & mAcc & OA \\
		\hline
		3DShapeNets~\cite{wu2015modelnet} & voxel & 77.3 & 84.7 \\
		VoxNet~\cite{maturana2015voxnet} & voxel & 83.0 & 85.9 \\
		Subvolume~\cite{qi2016volumetric} & voxel & 86.0 & 89.2 \\
		MVCNN~\cite{su15mvcnn} & image & -- & 90.1 \\
		PointNet~\cite{qi2017pointnet} & point & 86.2 & 89.2 \\
		A-SCN~\cite{xie2018attentional} & point & 87.6 & 90.0 \\
		Set Transformer~\cite{lee2019set} & point & -- & 90.4 \\
		PAT~\cite{yang2019modeling} & point & -- & 91.7 \\
		PointNet++~\cite{qi2017pointnet2} & point & -- & 91.9 \\
		SpecGCN~\cite{wang2018local} & point & -- & 92.1 \\
		PointCNN~\cite{li2018pointcnn} & point & 88.1 & 92.2 \\
		DGCNN~\cite{wang2019dgcnn} & point & 90.2 & 92.2 \\
		PointWeb~\cite{zhao2019pointweb} & point & 89.4 & 92.3 \\
		SpiderCNN~\cite{xu2018spidercnn} & point & -- & 92.4 \\
		PointConv~\cite{wu2019pointconv} & point & -- & 92.5 \\
		Point2Sequence~\cite{liu2019point2sequence} & point & 90.4 & 92.6 \\
		KPConv~\cite{thomas2019kpconv} & point & -- & 92.9 \\
		InterpCNN~\cite{mao2019interpolated} & point & -- & 93.0 \\
		\hline
		PointTransformer & point & \textbf{90.6} & \textbf{93.7} \\
		\bottomrule[1pt]
	\end{tabular}
	\caption{Shape classification results on the ModelNet40 dataset.}
	\label{tab:modelnet40result}
	\vspace{-2mm}
\end{table}

\subsection{Shape Classification}

\noindent\textbf{Data and metric.}
The ModelNet40~\cite{wu2015modelnet} dataset contains 12,311 CAD models with 40 object categories. They are split into 9,843 models for training and 2,468 for testing. We follow the data preparation procedure of Qi et al.~\cite{qi2017pointnet2} and uniformly sample the points from each CAD model together with the normal vectors from the object meshes. For evaluation metrics, we use the mean accuracy within each category (mAcc) and the overall accuracy (OA) over all classes.

\mypara{Performance comparison.}
The results are presented in Table~\ref{tab:modelnet40result}. The Point Transformer sets the new state of the art in both metrics. The overall accuracy of Point Transformer on ModelNet40 is 93.7\%. It outperforms strong graph-based models such as DGCNN~\cite{wang2019dgcnn}, attention-based models such as A-SCN~\cite{xie2018attentional} and Point2Sequence~\cite{liu2019point2sequence},
and strong point-based models such as KPConv~\cite{thomas2019kpconv}.

\begin{figure*}
	\centering
	\resizebox{0.97\linewidth}{!}{
	\begin{tabular}{c|cccccc}
	\includegraphics[width=0.16\linewidth]{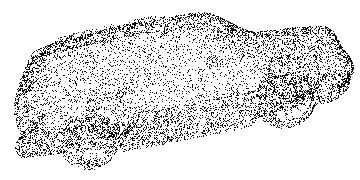}&
	\includegraphics[width=0.16\linewidth]{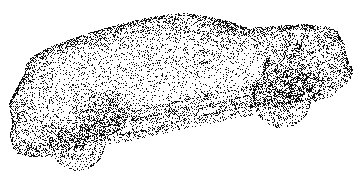}&
	\includegraphics[width=0.16\linewidth]{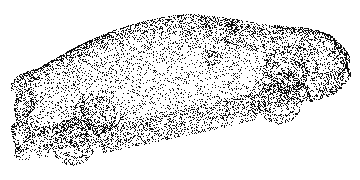}&
	\includegraphics[width=0.16\linewidth]{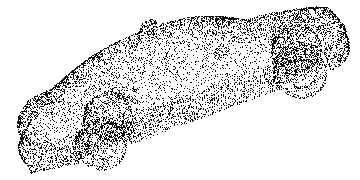}&
	\includegraphics[width=0.16\linewidth]{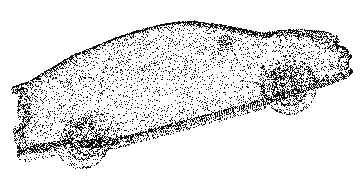}&
	\includegraphics[width=0.16\linewidth]{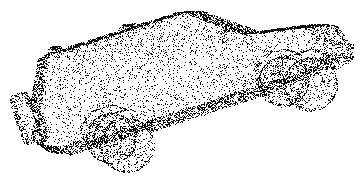}&
	\includegraphics[width=0.16\linewidth]{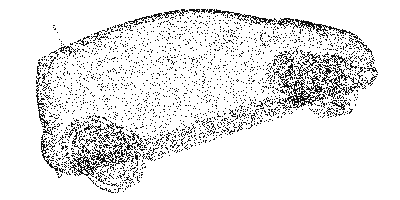}\\
	\includegraphics[width=0.16\linewidth]{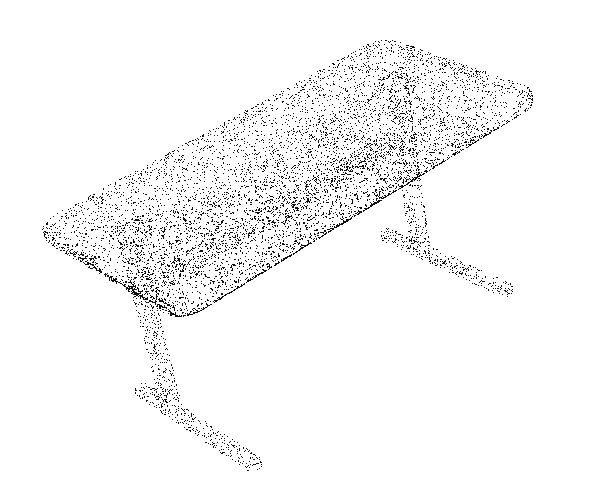}&
	\includegraphics[width=0.16\linewidth]{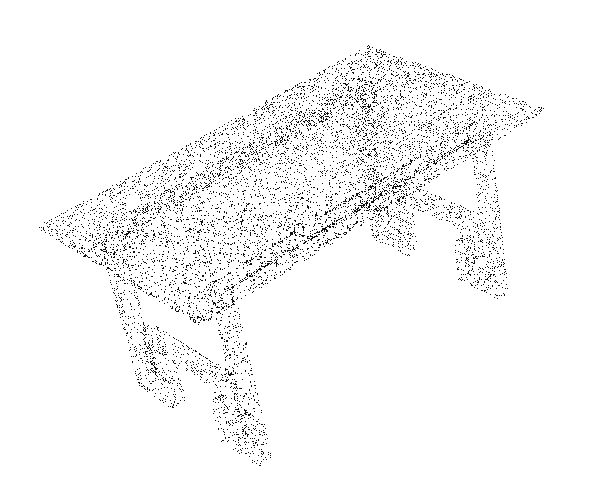}&
	\includegraphics[width=0.16\linewidth]{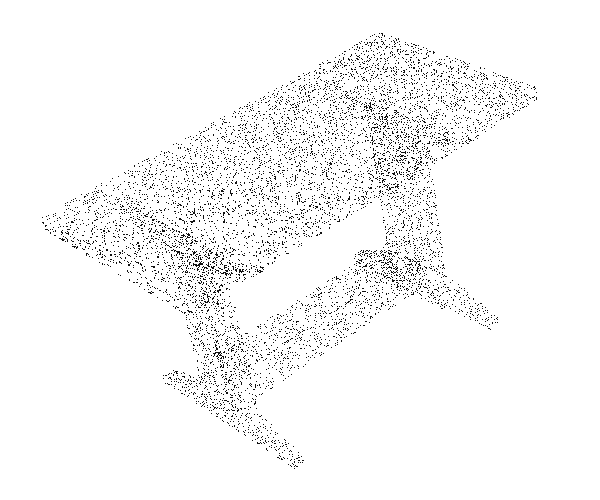}&
	\includegraphics[width=0.16\linewidth]{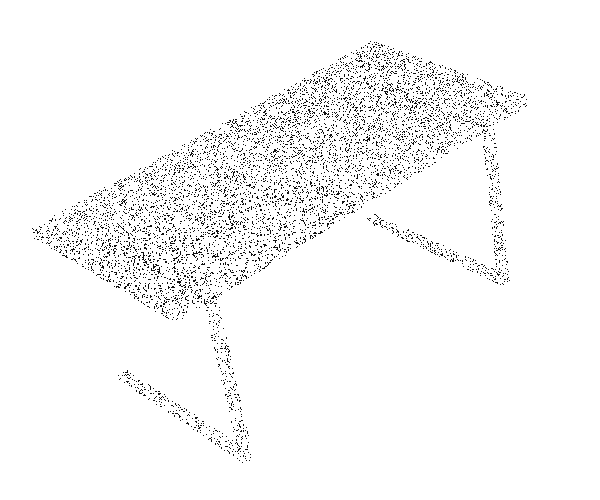}&
	\includegraphics[width=0.16\linewidth]{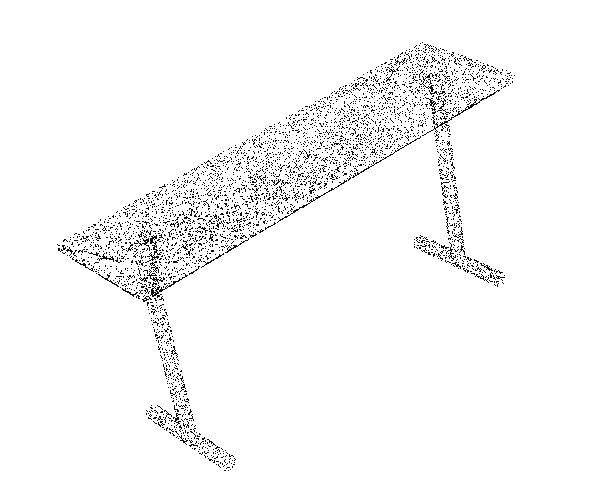}&
	\includegraphics[width=0.16\linewidth]{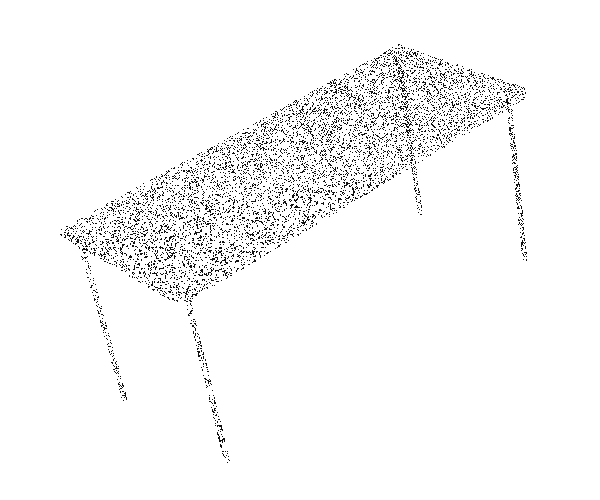}&
	\includegraphics[width=0.16\linewidth]{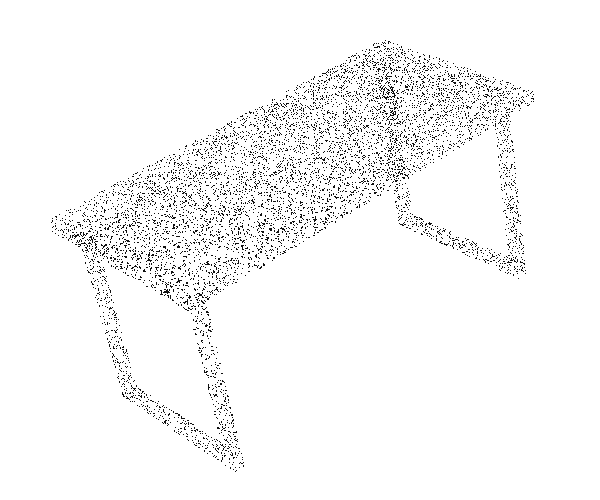}\\
	\end{tabular}}
	\caption{Visualization of shape retrieval results on the ModelNet40 dataset. The leftmost column shows the input query and the other columns show the retrieved models.}
	\label{fig:modelnet40visual}
\end{figure*}

\mypara{Visualization.}
To probe the representation learned by the Point Transformer, we conduct shape retrieval by retrieving nearest neighbors in the space of the output features produced by the Point Transformer. Some results are shown in Figure~\ref{fig:modelnet40visual}. The retrieved shapes are very similar to the query, and when they differ, they differ along aspects that we perceive as less semantically salient, such as legs of desks.

\begin{table}[t]
	\centering
	\begin{tabular}{ l | c c}
		\toprule[1pt]
		Method & cat.\ mIoU & ins.\ mIoU \\
		\hline
		PointNet~\cite{qi2017pointnet} & 80.4 & 83.7 \\
		A-SCN~\cite{xie2018attentional} & -- & 84.6 \\
		PCNN~\cite{wang2018pccn} & 81.8 & 85.1 \\
		PointNet++~\cite{qi2017pointnet2} & 81.9 & 85.1 \\
		DGCNN~\cite{wang2019dgcnn} & 82.3 & 85.1 \\
		Point2Sequence~\cite{liu2019point2sequence} & -- & 85.2 \\
		SpiderCNN~\cite{xu2018spidercnn}& 81.7 & 85.3 \\
		SPLATNet~\cite{su2018splatnet} & 83.7 & 85.4 \\
		PointConv~\cite{wu2019pointconv} & 82.8 & 85.7 \\
		SGPN~\cite{wang2018sgpn} & 82.8 & 85.8 \\
		PointCNN~\cite{li2018pointcnn} & 84.6 & 86.1 \\
		InterpCNN~\cite{mao2019interpolated} & 84.0 & 86.3 \\
		KPConv~\cite{thomas2019kpconv} & \textbf{85.1} & 86.4 \\
		\hline
		PointTransformer & 83.7 & \textbf{86.6} \\
		\bottomrule[1pt]
	\end{tabular}
	\caption{Object part segmentation results on the ShapeNetPart dataset.}
	\label{tab:shapenetpartresult}
	\vspace{-2mm}
\end{table}

\subsection{Object Part Segmentation}

\noindent\textbf{Data and metric.}
The ShapeNetPart dataset~\cite{yi2016scalable} is annotated for 3D object part segmentation. It consists of 16,880 models from 16 shape categories, with 14,006 3D models for training and 2,874 for testing. The number of parts for each category is between 2 and 6, with 50 different parts in total.
We use the sampled point sets produced by Qi et al.~\cite{qi2017pointnet2} for a fair comparison with prior work. For evaluation metrics, we report category mIoU and instance mIoU.

\mypara{Performance comparison.}
The results are presented in Table~\ref{tab:shapenetpartresult}. The Point Transformer outperforms all prior models as measured by instance mIoU.
(Note that we did not use loss-balancing during training, which can boost category mIoU.)

\begin{figure*}[!h]
	\centering
	\resizebox{0.97\linewidth}{!}{
	\begin{tabular}{ccccccccc}
	\includegraphics[width=0.195\linewidth]{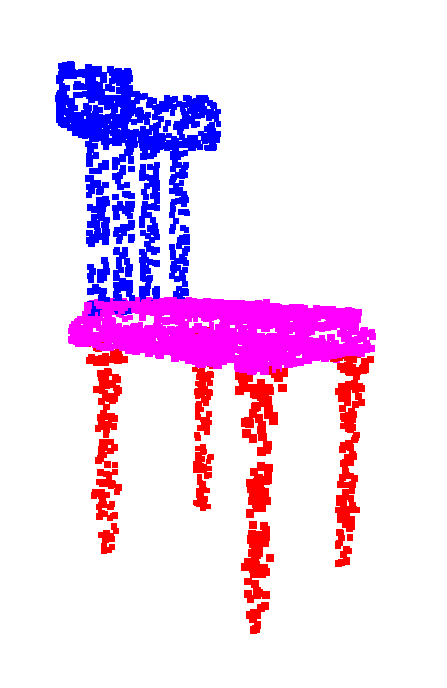}&
	\includegraphics[width=0.195\linewidth]{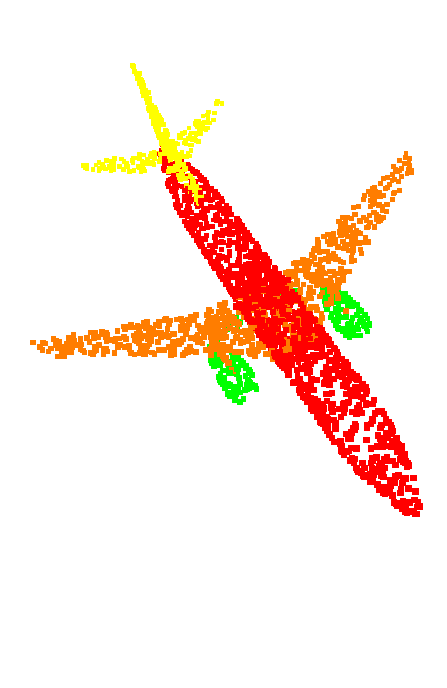}&
	\includegraphics[width=0.195\linewidth]{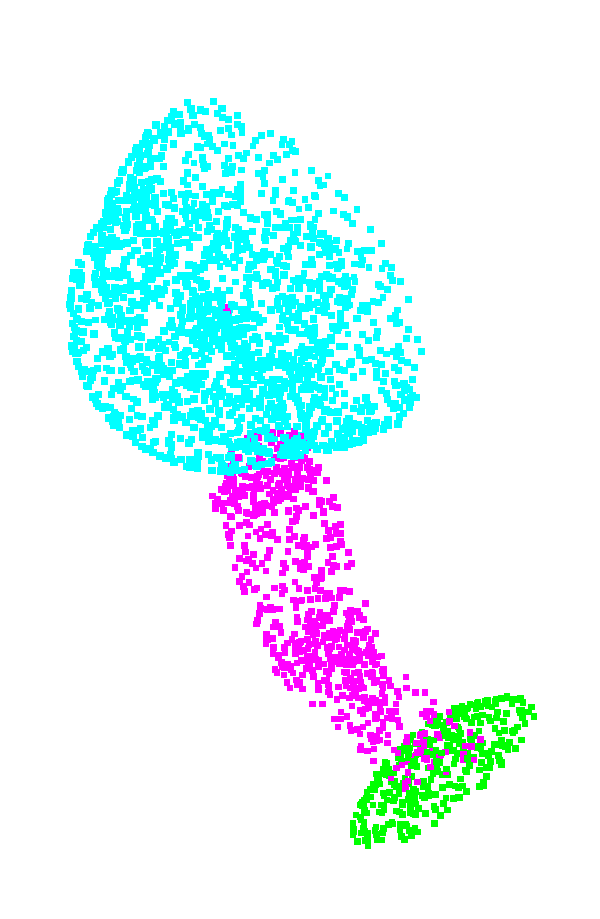}&
	\includegraphics[width=0.195\linewidth]{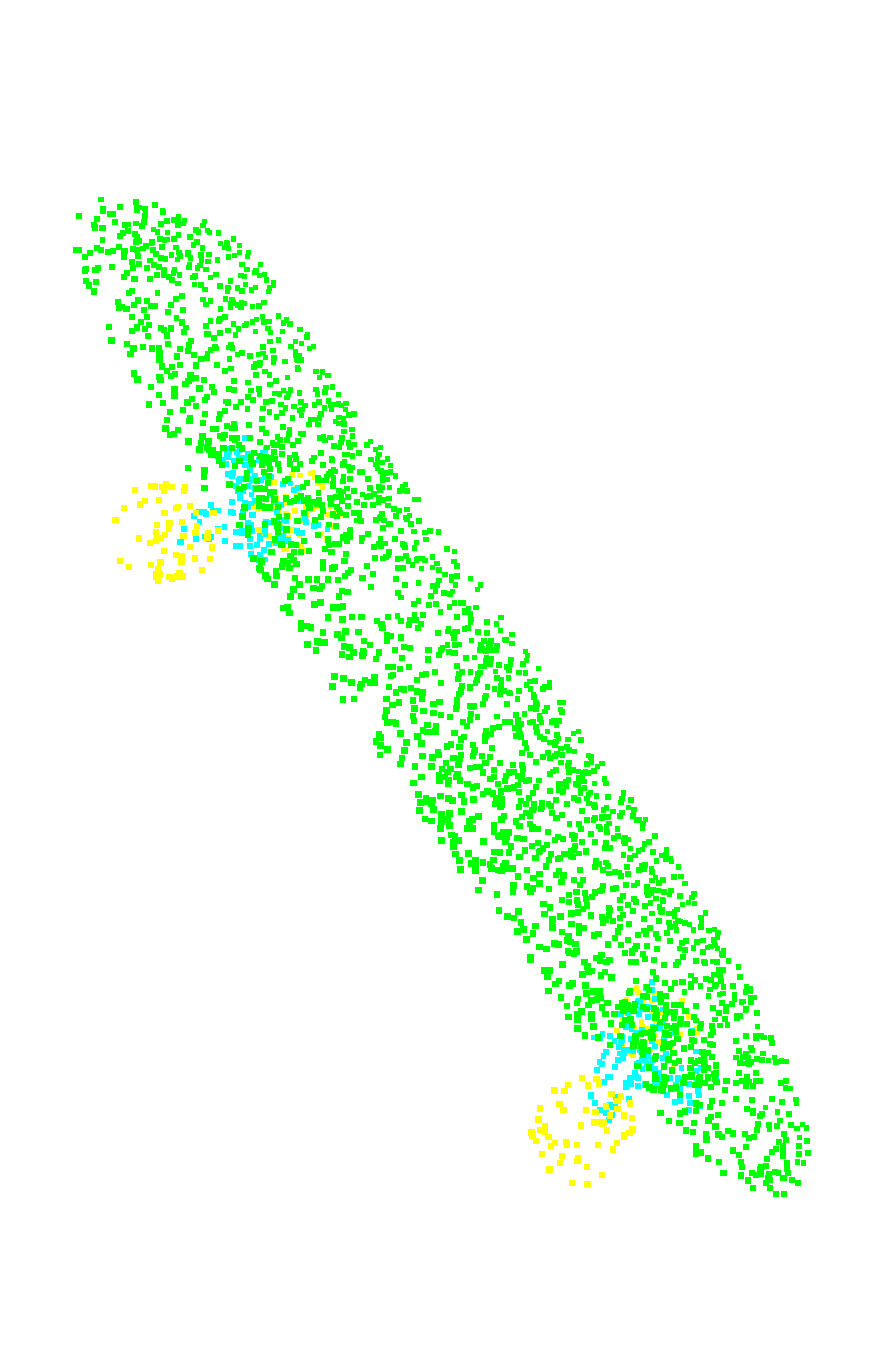}&
	\includegraphics[width=0.195\linewidth]{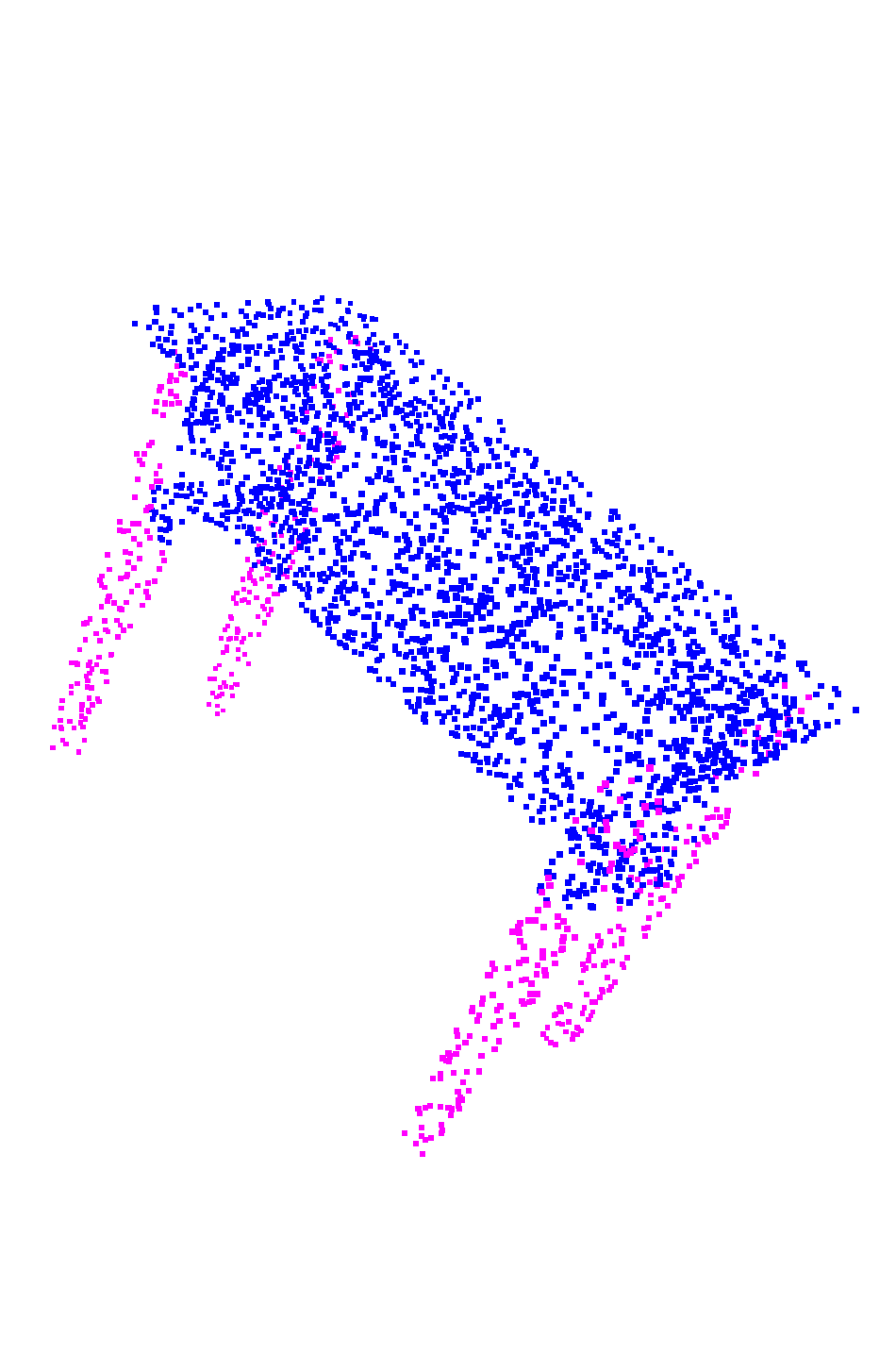}&
	\includegraphics[width=0.195\linewidth]{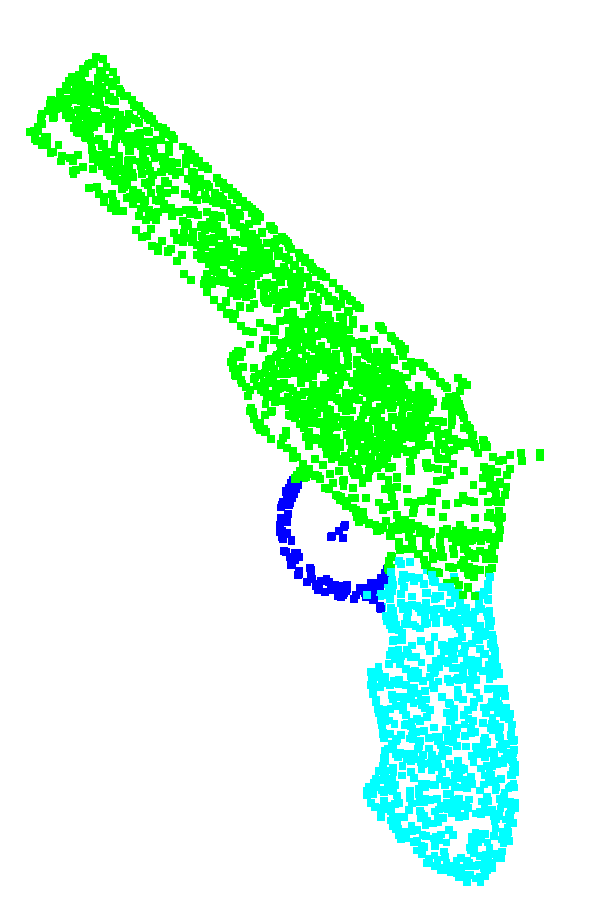}&
	\includegraphics[width=0.195\linewidth]{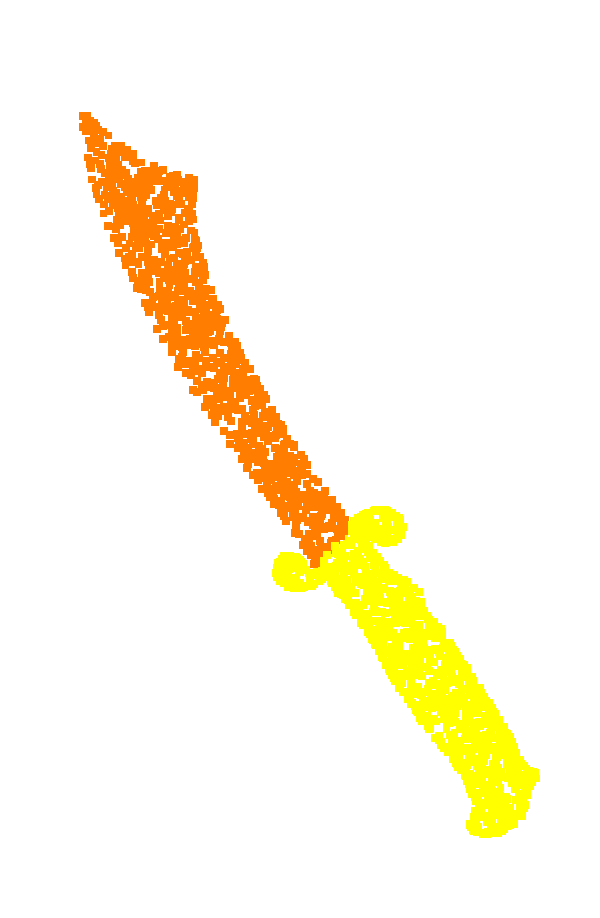}&
	\includegraphics[width=0.195\linewidth]{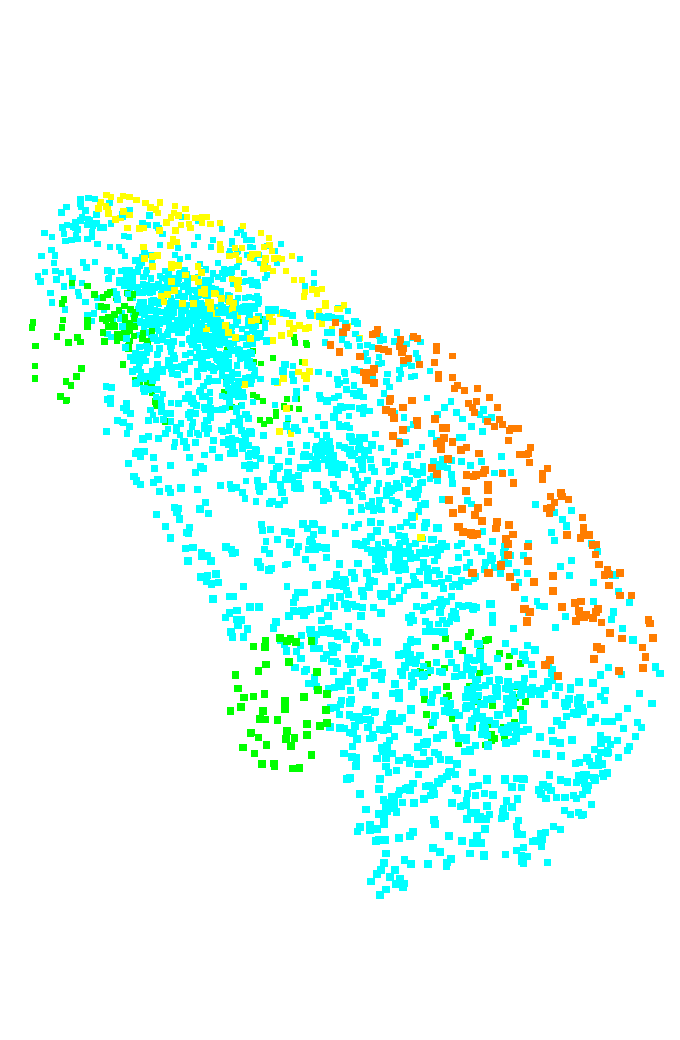}&
	\includegraphics[width=0.195\linewidth]{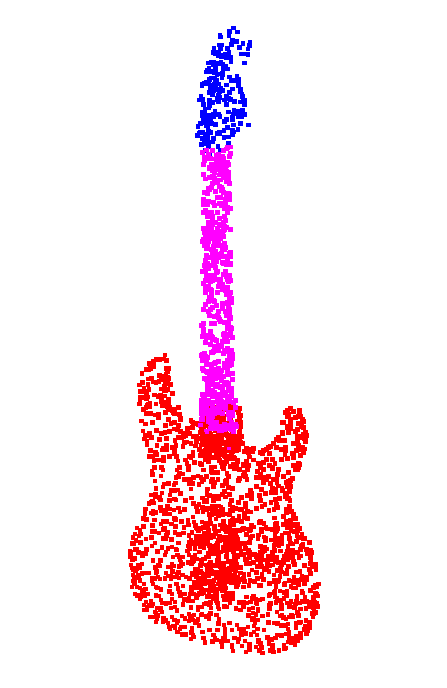}\\
	\includegraphics[width=0.195\linewidth]{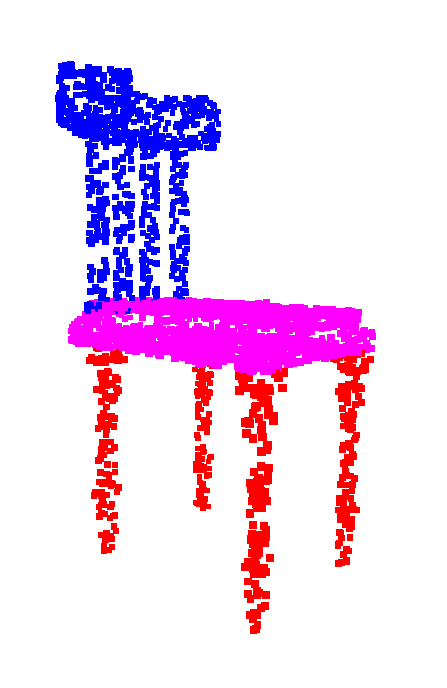}&
	\includegraphics[width=0.195\linewidth]{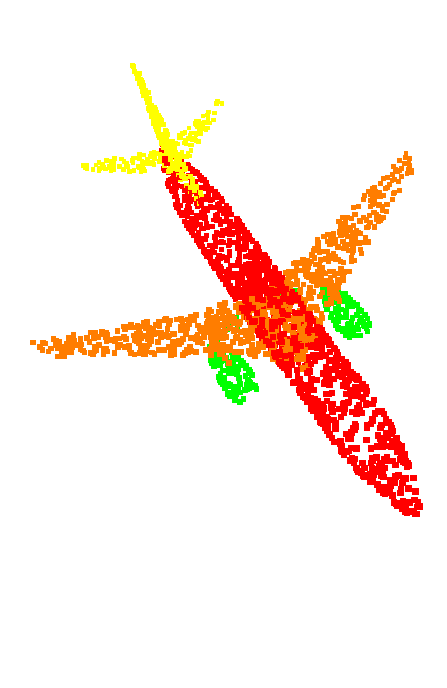}&
	\includegraphics[width=0.195\linewidth]{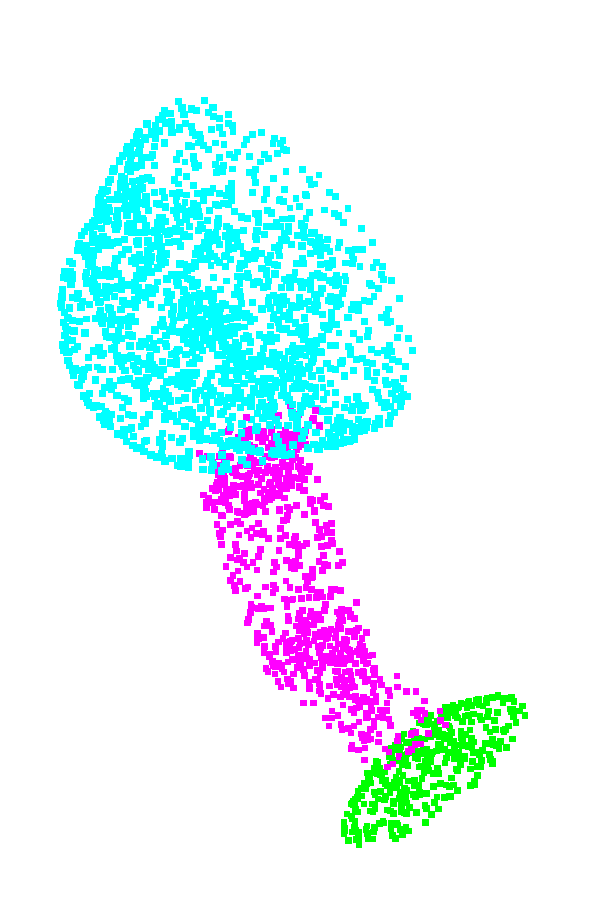}&
	\includegraphics[width=0.195\linewidth]{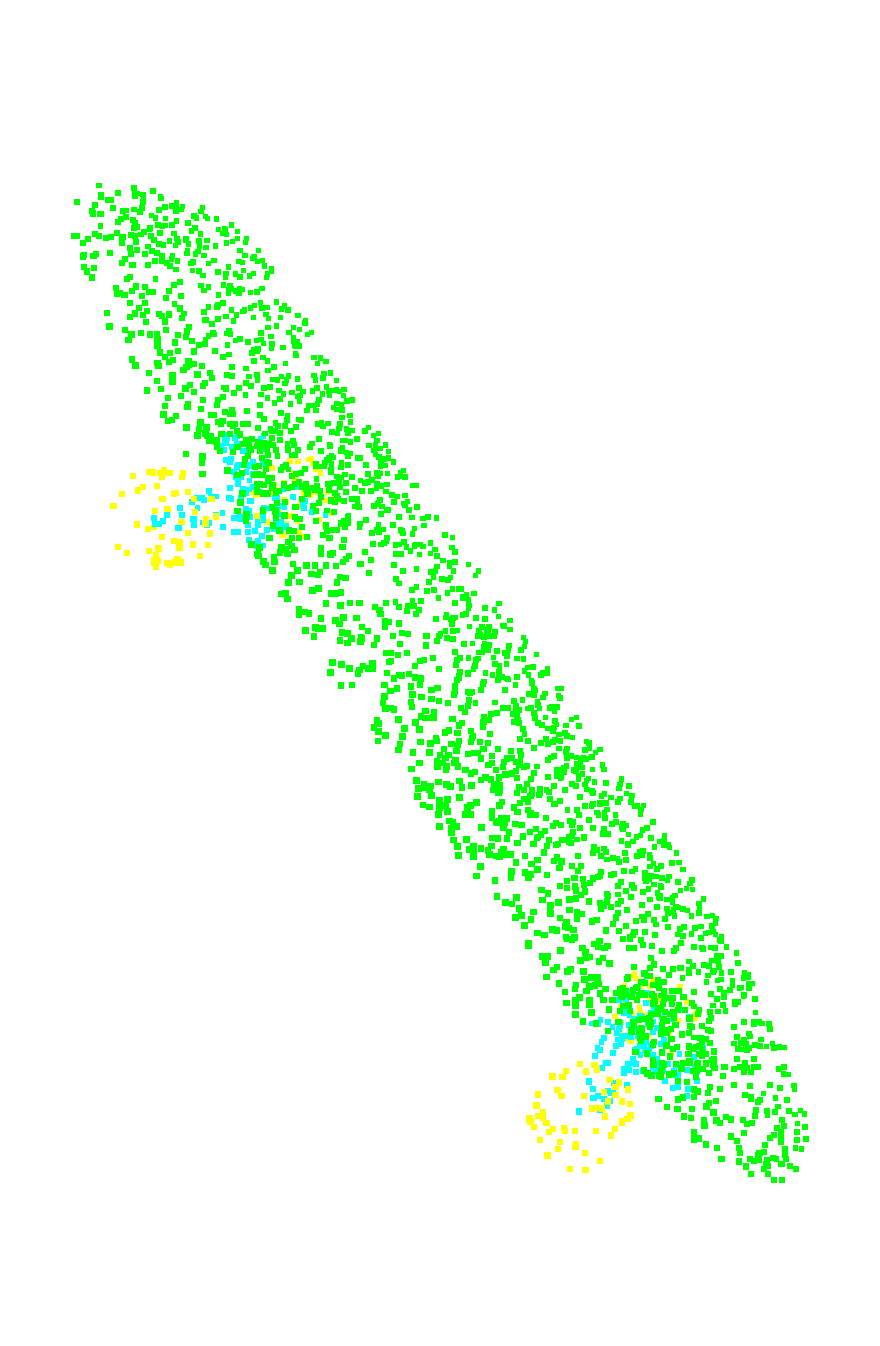}&
	\includegraphics[width=0.195\linewidth]{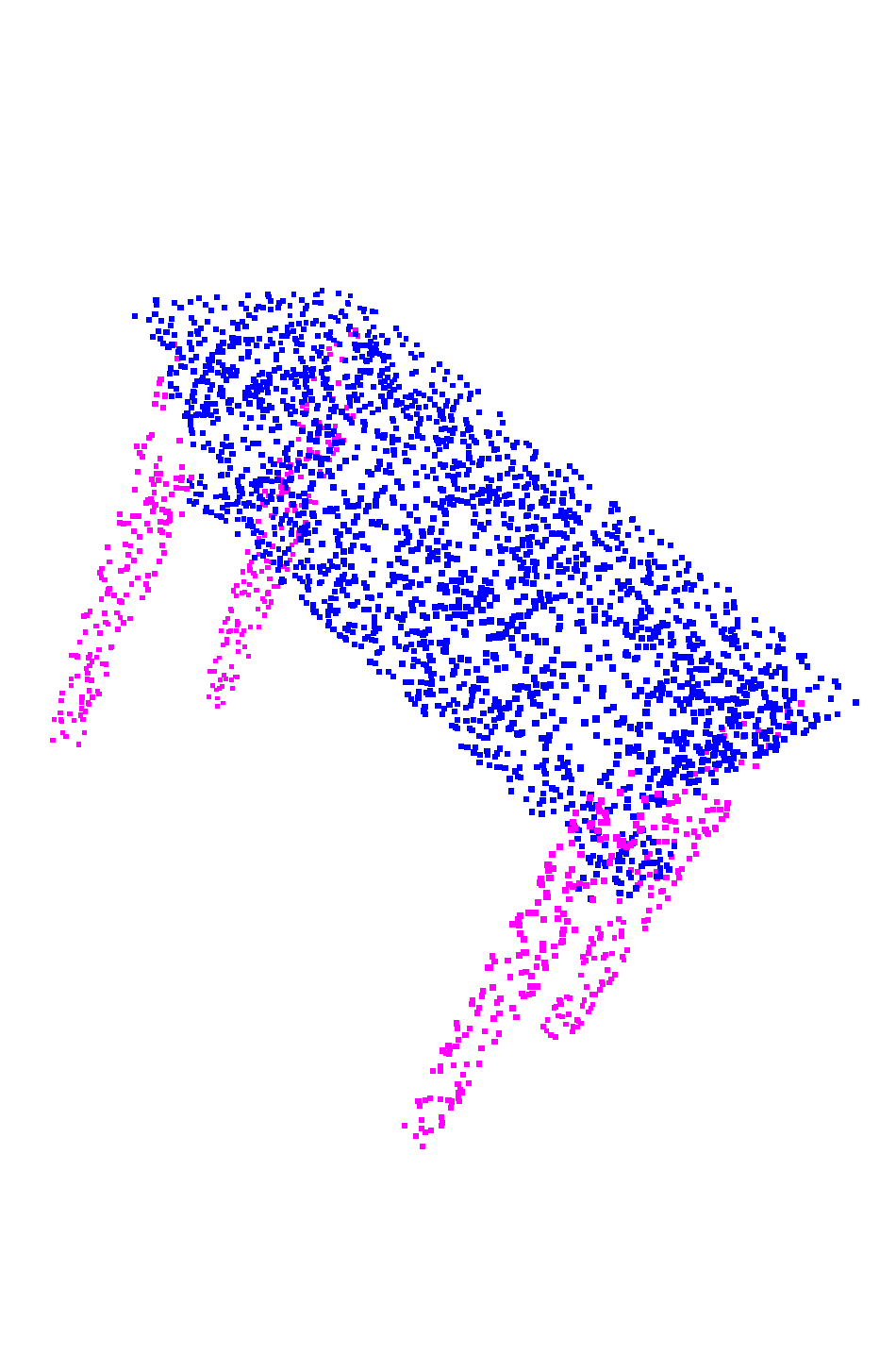}&
	\includegraphics[width=0.195\linewidth]{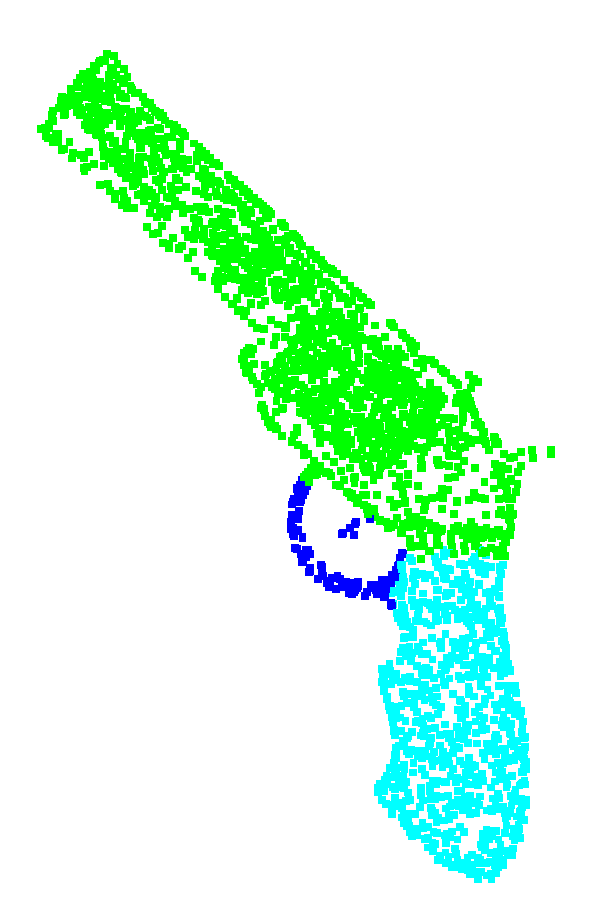}&
	\includegraphics[width=0.195\linewidth]{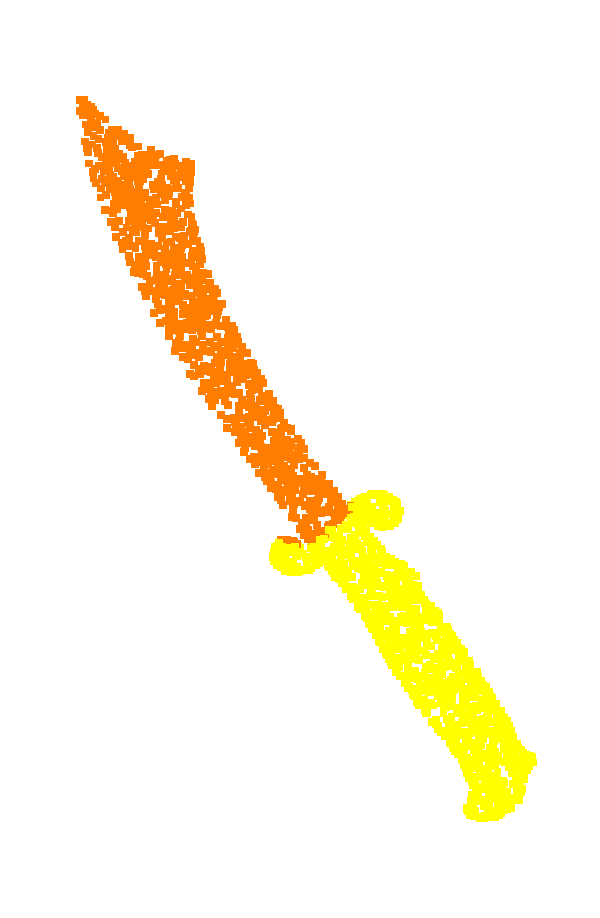}&
	\includegraphics[width=0.195\linewidth]{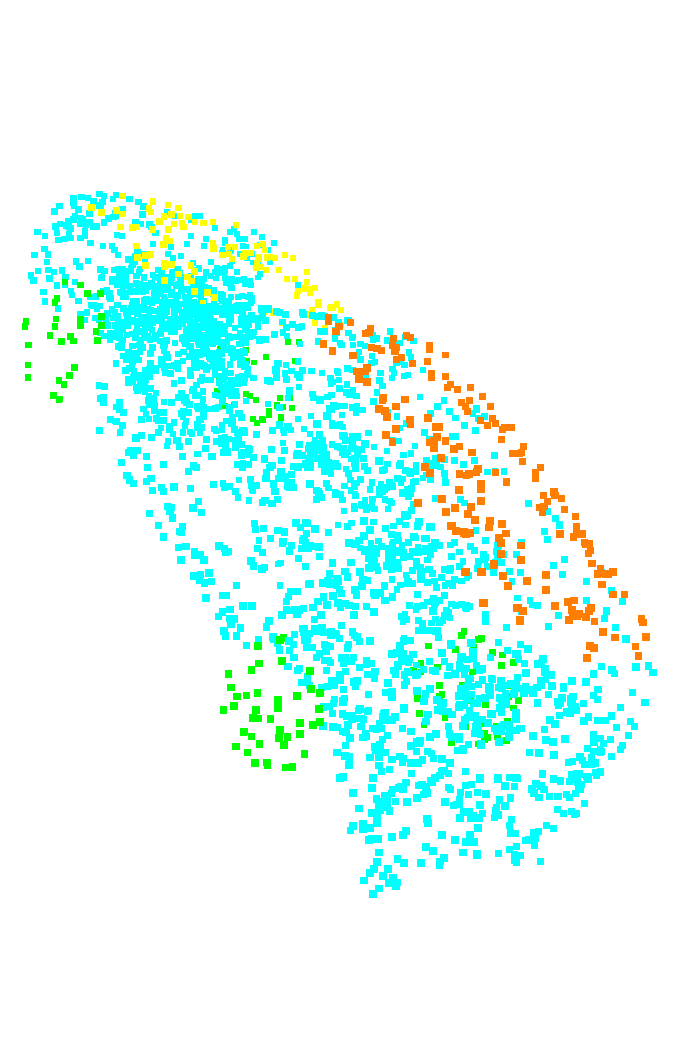}&
	\includegraphics[width=0.195\linewidth]{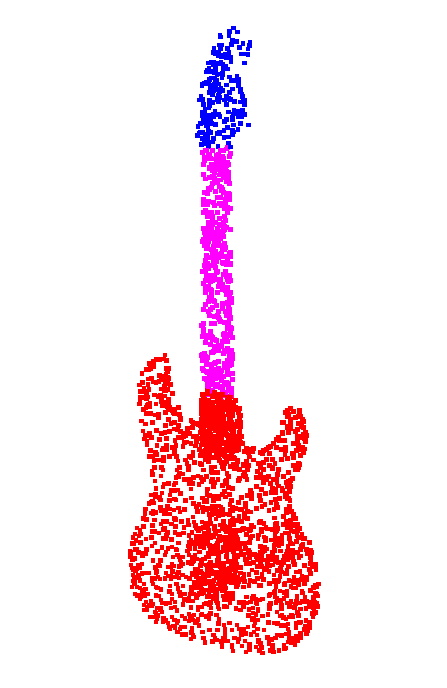}\\
	\end{tabular}}
	\caption{Visualization of object part segmentation results on the ShapeNetPart dataset. The ground truth is in the top row, Point Transformer predictions on the bottom.}
	\label{fig:shapenetpartvisual}
\end{figure*}

\mypara{Visualization.}
Object part segmentation results on a number of models are shown in Figure~\ref{fig:shapenetpartvisual}. The Point Transformer's part segmentation predictions are clean and close to the ground truth.

\begin{table}
	\centering
		\begin{tabular}{ c | c c c }
			\toprule[1pt]
			$k$ & mIoU & mAcc & OA \\
			\hline
			4 & 59.6 & 66.0 & 86.0 \\
			8 & 67.7 & 73.8 & 89.9 \\
			16 & \textbf{70.4} & \textbf{76.5} & \textbf{90.8} \\
			32 & 68.3 & 75.0 & 89.8 \\
			64 & 67.7 & 74.1 & 89.9 \\
			\bottomrule[1pt]
		\end{tabular}
	\caption{Ablation study: number of neighbors $k$ in the definition of local neighborhoods.}
	\label{tab:s3disknn}
\end{table}

\begin{table}
	\centering
		\begin{tabular}{ c | c c c }
			\toprule[1pt]
			Pos.\ encoding & mIoU & mAcc & OA \\
			\hline
			none & 64.6 & 71.9 & 88.2 \\
			absolute & 66.5 & 73.2 & 88.9 \\
			relative & \textbf{70.4} & \textbf{76.5} & \textbf{90.8} \\
			relative for attention & 67.0 & 73.0 & 89.3 \\
			relative for feature & 68.7 & 74.4 & 90.4 \\
			\bottomrule[1pt]
		\end{tabular}
	\caption{Ablation study: position encoding.}
	\label{tab:s3disposition}
\end{table}

\begin{table}
	\centering
		\begin{tabular}{ c | c c c }
			\toprule[1pt]
			Operator & mIoU & mAcc & OA \\
			\hline
			MLP & 61.7 & 68.6 & 87.1 \\
			MLP+pooling & 63.7 & 71.0 & 87.8 \\
			\hline
			scalar attention & 64.6 & 71.9 & 88.4 \\
			vector attention & \textbf{70.4} & \textbf{76.5} & \textbf{90.8} \\
			\bottomrule[1pt]
		\end{tabular}
	\caption{Ablation study: form of self-attention operator.}
	\label{tab:s3disattention}
\end{table}

\subsection{Ablation Study}
\label{sec:ablation}

We now conduct a number of controlled experiments that examine specific decisions in the Point Transformer design. These studies are performed on the semantic segmentation task on the S3DIS dataset, tested on Area 5.

\mypara{Number of neighbors.}
We first investigate the setting of the number of neighbors $k$, which is used in determining the local neighborhood around each point. The results are shown in Table~\ref{tab:s3disknn}. The best performance is achieved when $k$ is set to 16. When the neighborhood is smaller ($k=4$ or $k=8$), the model may not have sufficient context for its predictions. When the neighborhood is larger ($k=32$ or $k=64$), each self-attention layer is provided with a large number of datapoints, many of which may be farther and less relevant. This may introduce excessive noise into the processing, lowering the model's accuracy.

\mypara{Softmax regularization.}
We conduct an ablation study on the normalization function $\rho$ in Eq.~\ref{eq:pointtransformer}. The performance without softmax regularization on S3DIS Area5 is 66.5\%/72.8\%/89.3\%, in terms of mIoU/mAcc/OA. It is much lower than the performance with softmax regularization (70.4\%/76.5\%90.8\%). This suggests that the normalization is essential in this setting.

\mypara{Position encoding.}
We now study the choice of the position encoding $\delta$. The results are shown in Table~\ref{tab:s3disposition}. We can see that without position encoding, the performance drops significantly. With absolute position encoding, the performance is higher than without. Relative position encoding yields the highest performance. When relative position encoding is added only to the attention generation branch (first term in Eq.~\ref{eq:pointtransformer}) or only to the feature transformation branch (second term in Eq.~\ref{eq:pointtransformer}), the performance drops again, indicating that adding the relative position encoding to both branches is important.

\mypara{Attention type.}
Finally, we investigate the type of self-attention used in the point transformer layer. The results are shown in Table~\ref{tab:s3disattention}. We examine four conditions. `MLP' is a no-attention baseline that replaces the point transformer layer in the point transformer block with a pointwise MLP. `MLP+pooling' is a more advanced no-attention baseline that replaces the point transformer layer with a pointwise MLP followed by max pooling within each $k$NN neighborhood: this performs feature transformation at each point and enables each point to exchange information with its local neighborhood, but does not leverage attention mechanisms.
`scalar attention' replaces the vector attention used in Eq.~\ref{eq:pointtransformer} by scalar attention, as in Eq.~\ref{eq:scalar} and in the original transformer design~\cite{vaswani2017attention}. `vector attention' is the formulation we use, presented in Eq.~\ref{eq:pointtransformer}. We can see that scalar attention is more expressive than the no-attention baselines, but is in turn outperformed by vector attention. The performance gap between vector and scalar attention is significant: 70.4\% vs.\ 64.6\%, an improvement of 5.8 absolute percentage points. Vector attention is more expressive since it supports adaptive modulation of individual feature channels, not just whole feature vectors. This expressivity appears to be very beneficial in 3D data processing.

\section{Conclusion}
\label{sec:conclusion}
Transformers have revolutionized natural language processing and are making impressive gains in 2D image analysis. Inspired by this progress, we have developed a transformer architecture for 3D point clouds. Transformers are perhaps an even more natural fit for point cloud processing than they are for language or image processing, because point clouds are essentially sets embedded in a metric space, and the self-attention operator at the core of transformer networks is fundamentally a set operator. We have shown that beyond this conceptual compatibility, transformers are remarkably effective in point cloud processing, outperforming state-of-the-art designs from a variety of families: graph-based models, sparse convolutional networks, continuous convolutional networks, and others. We hope that our work will inspire further investigation of the properties of point transformers, the development of new operators and network designs, and the application of transformers to other tasks, such as 3D object detection.

\appendix
\section{Appendix}
\label{sec:appendix}
\renewcommand\thefigure{\thesection.\arabic{figure}}
\renewcommand\thetable{\thesection.\arabic{table}}
\setcounter{figure}{0} 
\setcounter{table}{0} 

\setcounter{table}{0}
\renewcommand{\thetable}{A.\arabic{table}}

\begin{table*}
	\centering
	\resizebox{1.0\linewidth}{!}{
		\begin{tabular}{ l | c c c | c c c c c c c c c c c c c}
			\toprule[1pt]
			Method & OA & mAcc & mIoU & ceiling & floor & wall & beam & column & window & door & table & chair & sofa & bookcase & board & clutter \\
			\hline
			PointNet~\cite{qi2017pointnet} & 78.5 & 66.2 & 47.6 & 88.0 & 88.7 & 69.3 & 42.4 & 23.1 & 47.5 & 51.6 & 54.1 & 42.0 & 9.6 & 38.2 & 29.4 & 35.2\\
			RSNet~\cite{huang2018recurrent} & -- & 66.5 & 56.5 & 92.5 & 92.8 & 78.6 & 32.8 & 34.4 & 51.6 & 68.1 & 60.1 & 59.7 & 50.2 & 16.4 & 44.9 & 52.0\\
			SPGraph~\cite{landrieu2018spg} & 85.5 & 73.0 & 62.1 & 89.9 & 95.1 & 76.4 & 62.8 & 47.1 & 55.3 & 68.4 & 73.5 & 69.2 & 63.2 & 45.9 & 8.7 & 52.9\\
			PAT~\cite{yang2019modeling} & -- & 76.5 & 64.3 & 93.0 & 98.4 & 73.5 & 58.5 & 38.9 & 77.4 & 67.7 & 62.7 & 67.3 & 30.6 & 59.6 & 66.6 & 41.4\\
			PointCNN~\cite{li2018pointcnn} & 88.1 & 75.6 & 65.4 & 94.8 & 97.3 & 75.8 & 63.3 & 51.7 & 58.4 & 57.2 & 71.6 & 69.1 & 39.1 & 61.2 & 52.2 & 58.6\\
			PointWeb~\cite{zhao2019pointweb} & 87.3 & 76.2 & 66.7 & 93.5 & 94.2 & 80.8 & 52.4 & 41.3 & 64.9 & 68.1 & 71.4 & 67.1 & 50.3 & 62.7 & 62.2 & 58.5 \\
			ShellNet~\cite{zhang2019shellnet} & 87.1 & -- & 66.8 & 90.2 & 93.6 & 79.9 & 60.4 & 44.1 & 64.9 & 52.9 & 71.6 & 84.7 & 53.8 & 64.6 & 48.6 & 59.4 \\
			RandLA-Net~\cite{thomas2019kpconv} & 88.0 & 82.0 & 70.0 & 93.1 & 96.1 & 80.6 & 62.4 & 48.0 & 64.4 & 69.4 & 69.4 & 76.4 & 60.0 & 64.2 & 65.9 & 60.1\\
			KPConv~\cite{thomas2019kpconv} & -- & 79.1 & 70.6 & 93.6 & 92.4 & 83.1 & 63.9 & 54.3 & 66.1 & 76.6 & 64.0 & 57.8 & 74.9 & 69.3 & 61.3 & 60.3 \\
			\hline
			PointTransformer & \textbf{90.2} & \textbf{81.9} & \textbf{73.5} & 94.3 & 97.5 & 84.7 & 55.6 & 58.1 & 66.1 & 78.2 & 77.6 & 74.1 & 67.3 & 71.2 & 65.7 & 64.8 \\
			\bottomrule[1pt]
		\end{tabular}}
	\caption{Semantic segmentation results on S3DIS, evaluated with 6-fold cross-validation.}
	\label{tab:s3disresult3}
\end{table*}

\mypara{More detailed results.}
In Table~\ref{tab:s3disresult3}, we present the detailed comparison of the semantic segmentation results on the S3DIS dataset~\cite{armeni2016s3dis}, under the 6-fold cross-validation setting. We get the highest mIoU as 73.5\%, outperforming previous approaches (e.g., RandLA-Net~\cite{hu2020randla} and KPConv~\cite{thomas2019kpconv}) by a large margin. For most of the categories (like wall, column, table, etc.), our method gets the best accuracy. We will release all the implementation details and trained models to the community soon.

\mypara{Inference time and memory.}
We test the inference time and memory consumption of Point Transformer on one Quadro RTX 6000, with different size of input point clouds. The inference time and memory consumption are 44ms/86ms/222ms/719ms and 1702M/2064M/2800M/4266M for 10k/20k/40k/80k input points respectively and they can be further reduced with optimized implementation.

\mypara{$k$NN efficiency.}
For $k$NN, when constructing local point cloud regions, previous methods like KPConv~\cite{thomas2019kpconv} and RandLA-Net~\cite{hu2020randla} use precomputed $k$NN indices, which limits the flexibility of the overall framework. In our architecture, we implement a high-efficiency solution for $k$NN using the heap sort algorithm. We test the running time of our efficient implementation on one Quadro RTX 6000; the results are listed in Table~\ref{tab:knn}. We also test some naive implementations, the running time is 56ms/228ms when given 10k/20k points, which is much slower than ours. Moreover, naive implementations run out of memory when given larger point clouds.

\begin{table}
	\centering
	\begin{tabular}{ c | c c c c c c}
		\toprule[1pt]
		\#pts & $k$=8 & $k$=16 & $k$=32 & $k$=64 & $k$=128 & $k$=256 \\
		\hline
		10k & 2 & 2 & 5  & 10 & 17 & 21 \\
		20k & 3 & 5 & 8  & 23 & 43 & 49 \\
		40k & 8 & 12 & 26  & 71 & 127 & 144 \\
		80k & 23 & 37 & 82  & 198 & 356 & 399 \\
		100k & 32 & 46 & 99 & 248 & 445 & 494 \\
		200k & 104 & 125 & 225 & 545 & 992 & 1091 \\
		500k & 639 & 695 & 867 & 1589 & 2865 & 3143 \\
		1m & 2496 & 2648 & 2949 & 4087 & 6362 & 6878 \\
		\bottomrule[1pt]
	\end{tabular}
	\caption{High-efficiency $k$NN implementation with heap sort algorithm. The leftmost column stands for the number of points and the topmost row specifies the number of nearest neighbors. The reported running time is in milliseconds.}
	\label{tab:knn}
\end{table}

\balance

{\small
	\bibliographystyle{ieee_fullname}
	\bibliography{paper}
}

\end{document}